\documentclass[10pt,twocolumn,letterpaper]{article}

\usepackage{cvpr}
\usepackage{times}
\usepackage{epsfig}
\usepackage{graphicx}
\usepackage{amsmath}
\usepackage{amssymb}

\usepackage{caption}
\usepackage{subcaption}
\usepackage[export]{adjustbox} %
\usepackage{booktabs}
\usepackage{arydshln}
\usepackage{dashrule}

\usepackage{bigstrut}
\usepackage{multirow}
\usepackage{hhline}

\usepackage{pifont}
\newcommand{\cmark}{\ding{51}}%
\newcommand{\xmark}{\ding{55}}%

\usepackage[numbers,sort]{natbib}
\usepackage{placeins} 

\renewcommand{\paragraph}[1]{\medskip\noindent{\bf{#1}}}

\newif\ifkeepComments
\keepCommentstrue

\graphicspath{{images/}{supp_mat/}}

\cvprfinalcopy

\newif\ifArXivVersion
\ArXivVersiontrue

\newcommand{\isArXiv}[2]{\ifArXivVersion #1\else #2 \fi}

\usepackage[pagebackref=true,breaklinks=true,letterpaper=true,colorlinks,bookmarks=false]{hyperref}
\newif\ifbackrefshowonlyfirst
\backrefshowonlyfirsttrue
\makeatletter
\let\BR@direct@old@hyper@natlinkstart\hyper@natlinkstart
\renewcommand*{\hyper@natlinkstart}{\phantomsection\BR@direct@old@hyper@natlinkstart}
\let\BR@direct@oldBR@citex\BR@citex
\renewcommand*{\BR@citex}{\phantomsection\BR@direct@oldBR@citex}%

\long\def\hyper@page@BR@direct@ref#1#2#3{\hyperlink{#3}{#1}}

\ifx\backrefxxx\hyper@page@backref
    \let\backrefxxx\hyper@page@BR@direct@ref
    \ifbackrefshowonlyfirst
    \fi
\else
    \ifbackrefshowonlyfirst
    \fi
\fi

\RequirePackage{etoolbox}
\patchcmd{\Hy@backout}{Doc-Start}{\@currentHref}{}{\errmessage{I can't seem to patch backref}}
\makeatother
\hypersetup{pdfauthor={Ignacio Rocco},pdftitle={End-to-end weakly-supervised semantic alignment}}
\makeatletter
\renewcommand*{\@fnsymbol}[1]{\ensuremath{\ifcase#1\or \or \or \or
\or \or \or \or \or \else\@ctrerr\fi}}
\makeatother

\def\cvprPaperID{1016} 

\usepackage[utf8]{inputenc}

\begin{document}

\title{End-to-end weakly-supervised semantic alignment}

\author{Ignacio Rocco$^{1,2}$ \qquad Relja Arandjelovi\'{c}\,$^{3}$
\qquad Josef Sivic$^{1,2,4}$\\
$^1$DI ENS \quad \quad \quad  \quad  $^2$Inria  \quad \quad \quad \quad $^3$DeepMind \quad \quad  \quad \quad $^4$CIIRC, CTU in Prague
\thanks{$^1$Département d’informatique de l’ENS, École normale supérieure, CNRS, PSL Research University, 75005 Paris, France.}
\thanks{$^4$Czech Institute of Informatics, Robotics and Cybernetics at the Czech Technical University in Prague.}
}

\maketitle

\global\csname @topnum\endcsname 0
\global\csname @botnum\endcsname 0

\begin{abstract}

We tackle the task of semantic alignment where the goal is to compute dense semantic correspondence 
aligning two images depicting objects of the same category. This is a challenging task due to large intra-class variation, changes in viewpoint and background clutter.   We present the following three principal contributions. 
First, we develop a convolutional neural network architecture for semantic alignment  that is trainable in an end-to-end manner from weak image-level supervision in the form of matching image pairs. The outcome is that parameters are learnt from rich appearance variation present in different but semantically related images without the need for tedious manual annotation of correspondences at training time.
Second, the main component of this architecture is a differentiable soft inlier scoring module, inspired by the RANSAC inlier scoring procedure, that computes the quality of the alignment based on only geometrically consistent correspondences thereby reducing the effect of background clutter. 
Third, we demonstrate that the proposed approach achieves state-of-the-art performance on multiple standard benchmarks for semantic alignment.    

\end{abstract}

\begin{figure}[t]
\centering
\includegraphics[width=\columnwidth]{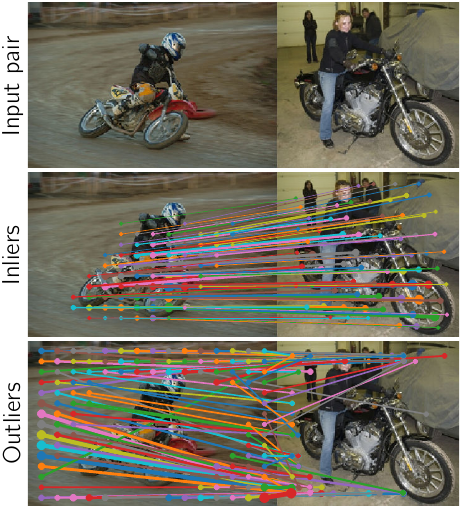} 
\captionsetup{font={small}}
\caption{ 
We describe a CNN architecture that, given an input image pair (top), 
outputs dense semantic correspondence between the two images together with the aligning geometric transformation  (middle) and discards  geometrically inconsistent matches (bottom). The alignment model is learnt from weak supervision in the form of matching image pairs without correspondences.}\label{fig:teaser}
\end{figure}

\section{Introduction\label{sec:intro}}

Finding correspondence is one of the fundamental problems in computer vision. Initial work has focused on finding correspondence between images depicting the same object or scene with applications in image stitching \cite{szeliski2006image}, multi-view 3D reconstruction~\cite{hartley2003multiple}, motion estimation \cite{weinzaepfel2013deepflow,fischer2015flownet} or tracking \cite{newcombe2011dtam,engel2014lsd}.  
In this work we study the problem of finding category-level correspondence, or {\em semantic alignment}~\cite{Berg05,liu2008sift}, where the goal is to establish dense correspondence between different objects belonging to the same category, such as the two different motorcycles  illustrated in Fig.~\ref{fig:teaser}. This is an important problem with applications in object recognition~\cite{liu2011sift}, image editing~\cite{Dale09}, or robotics~\cite{Nikandrova15}.
This is also an extremely challenging task because of the large intra-class variation, changes in viewpoint and presence of background clutter.

The current best semantic alignment methods~\cite{kim2017dctm,han2017scnet,novotny2017} employ powerful image representations based on convolutional neural networks coupled with a geometric deformation model. However, these methods suffer from one of the following two major limitations.
First, the image representation and the geometric alignment model are not trained together in an end-to-end manner. Typically, the image representation is trained on some auxiliary task such as image classification and then employed in an often ad-hoc geometric alignment model. 
Second, while trainable geometric alignment models exist~\cite{Rocco17,Brachmann17}, they require strong supervision in the form of ground truth correspondences, which is hard to obtain for a diverse set of real images on a large scale. 

In this paper, we address both these limitations and develop a semantic alignment model that is {\em trainable end-to-end} from {\em weakly supervised} data in the form of matching
image pairs
without the need for ground truth correspondences. 
To achieve that we design a novel convolutional neural network architecture for semantic alignment with a differentiable soft inlier scoring module inspired by the RANSAC inlier scoring procedure. The resulting architecture is end-to-end trainable with only image-level supervision.  The outcome is that the image representation can be trained from rich appearance variations present in different but semantically related image pairs, rather than synthetically deformed imagery~\cite{kanazawa2016warpnet,Rocco17}. We show that our approach allows to significantly improve the performance of the baseline deep CNN alignment model, achieving state-of-the-art performance on multiple standard benchmarks for semantic alignment. 
Our code and trained models are available online~\cite{webpage}.

\begin{figure*}
\includegraphics[width=\textwidth]{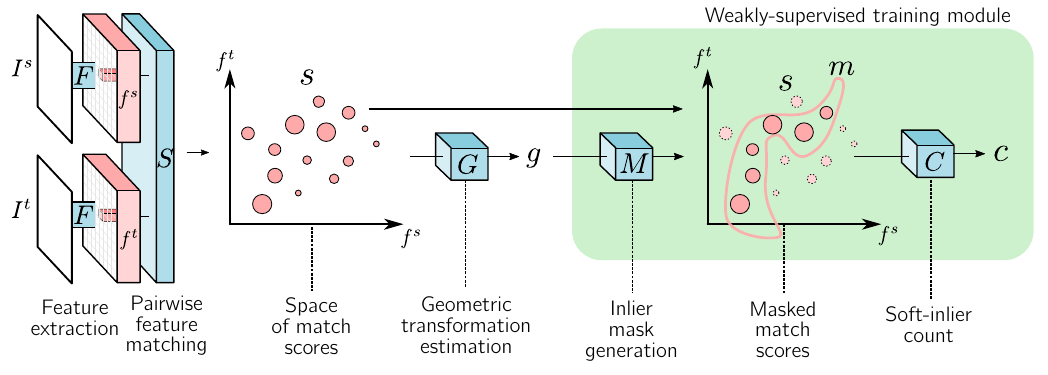} 
\captionsetup{font={small}}
\caption{{\bf End-to-end weakly-supervised alignment.}
Source and target images $(I^s,I^t)$ are passed through an alignment network
used to estimate the geometric transformation $g$.
Then, the soft-inlier count is computed (in green) by first finding the inlier region $m$ 
in agreement with $g$, and then adding up the pairwise matching scores inside this
area. The soft-inlier count is differentiable, which allows the whole model to be trained using back-propagation. Functions are represented in blue and tensors in pink.} \label{fig:proposed-arch}
\end{figure*}

\section{Related work}
The problem of semantic alignment has received significant attention in the last few years with progress in both (i) image descriptors and (ii) geometric models. The key innovation has been making the two components trainable from data.  
We summarize the recent progress in~Table~\ref{tab:rel_work} where we indicate for each method  whether the descriptor (D) or the alignment model (A) are trainable, whether the entire architecture is trainable end-to-end (E-E), and whether the required supervision is strong (s) or weak (w). 

\begin{table}[t!]
\small
\begin{center}
\begin{tabular}{@{\hskip 1mm}l@{\hskip 2mm}l@{\hskip 2mm}l@{\hskip 2mm}c@{\hskip 2mm}c@{\hskip 1mm}c@{\hskip 1mm}c@{\hskip 1mm}}
\toprule
\multirow{2}{*}{Paper} &	\multirow{2}{*}{Descriptor} &	\multirow{2}{*}{\begin{tabular}[x]{@{}c@{}}Alignment\\method\end{tabular}} &	\multicolumn{4}{c}{Trainable}  \\
& & & D & A & E-E & S\\
\midrule\midrule
Liu \emph{et al.}`11~\cite{liu2011sift} &	SIFT &	SIFT Flow &	\xmark &	\xmark &	\xmark & -\\
\midrule
Kim \emph{et al.}`13~\cite{Kim13} &	SIFT+PCA &	DSP &	\xmark &	\xmark &	\xmark & -\\
\midrule
Taniai \emph{et al.}`16~\cite{taniai2016joint} &	HOG &	TSS &	\xmark &	\xmark &	\xmark & -\\
\midrule
Ham \emph{et al.}`16~\cite{ham2017} & 	HOG &	PF-LOM &	\xmark &	\xmark &	\xmark & -\\
\midrule
Yang \emph{et al.}`17~\cite{yang2017object} &	HOG &	OADSC &	\xmark &	\xmark &	\xmark & -\\
\midrule
Ufer \emph{et al.}`17~\cite{ufer2017deep} &	AlexNet &	DSFM &	\xmark &	\xmark &	\xmark & -\\
\midrule
\midrule
\multirow{2}{*}{\begin{tabular}[x]{@{}l@{}}Novotny \emph{et al.}`17~\cite{novotny2017}\end{tabular}} &	\multirow{2}{*}{AnchorNet} &	DSP &	\cmark &	\xmark &	\xmark & w\\
 &	 &	PF-LOM &	\cmark &	\xmark &	\xmark & w\\
 \midrule
\multirow{2}{*}{Kim \emph{et al.}`17~\cite{kim2017fcss}} &	\multirow{2}{*}{FCSS} &	SIFT Flow &	\cmark &	\xmark &	\xmark & s\\
 &	 &	PF-LOM &	\cmark &	\xmark &	\xmark & s\\
 \midrule
Kim \emph{et al.}`17~\cite{kim2017dctm} &	FCSS &	DCTM &	\cmark &	\xmark & 	\xmark & s\\
\midrule
\midrule
\multirow{3}{*}{Han \emph{et al.}`17~\cite{han2017scnet}}&	\multirow{3}{*}{VGG-16} &	SCNet-A &	\cmark &	\cmark &	\xmark & s\\
&	 &	SCNet-AG &	\cmark &	\cmark &	\xmark & s\\
&	 &	SCNet-AG+ &	\cmark &	\cmark &	\xmark & s\\
\midrule
\midrule
\multirow{2}{*}{Rocco \emph{et al.}`17~\cite{Rocco17}} &	VGG-16 &	CNN Geo. &	\cmark &	\cmark &	\cmark & s\\
 &	ResNet-101 &	CNN Geo. &	\cmark &	\cmark &	\cmark & s\\
 \midrule
Proposed method &	ResNet-101 &	CNN Geo. &	\cmark &	\cmark &	\cmark & w\\
\bottomrule
\end{tabular}
\captionsetup{font={small}}
\vspace*{-1mm}
\caption{{\bf Comparison of recent related work.} The table indicates employed image descriptor and alignment method. The last four columns show which components of the approach are trained for the semantic alignment task:  descriptor (D), alignment (A) or both in end-to-end manner (E-E); and the level of supervision (S): strong (s) or weak (w).}
\label{tab:rel_work}
\vspace*{-5mm}
\end{center}
\end{table}

Early methods, such as \cite{Berg05,liu2011sift,Kim13}, employed hand-engineered descriptors like SIFT or HOG together with hand-engineered alignment models based on minimizing a given matching energy. This approach has been quite successful~\cite{ham2017,taniai2016joint,yang2017object,ufer2017deep} using in some cases~\cite{ufer2017deep} pre-trained (but fixed) convolutional neural network (CNN) descriptors. However, none of these methods train the image descriptor or the geometric model directly for semantic  alignment.

Others~\cite{novotny2017,kim2017fcss,kim2017dctm} have investigated trainable image descriptors for semantic matching
but have combined them with hand-engineered alignment models still rendering the alignment pipeline not trainable end-to-end.

Finally, recent work~\cite{han2017scnet,Rocco17} has employed trainable CNN descriptors together with trainable geometric alignment methods. However, in~\cite{han2017scnet} the matching is learned at the object-proposal level and a non-trainable fusion step is necessary to output the final alignment making the method non end-to-end trainable. On the contrary, \cite{Rocco17} estimate a parametric geometric model, which can be converted into dense pixel correspondences in a differentiable way, making the method end-to-end trainable. However, the method is trained with strong supervision in the form of ground truth correspondences obtained from synthetically warped images, which significantly limits the appearance variation in the training data.

\paragraph{Contributions.}
We develop a network architecture where both the descriptor and the alignment model are trainable in an end-to-end manner from weakly supervised data. This enables training from real images with rich appearance variation and without the need for manual ground-truth correspondence. We demonstrate that the proposed approach significantly improves alignment results achieving state-of-the-art performance on several datasets for semantic alignment. 

\section{Weakly-supervised semantic alignment}

This section presents a method for training a semantic alignment model in an
end-to-end fashion using only weak supervision -- the information that two
images should match -- but without access to the underlying geometric transformation
at training time.
The approach is outlined in Fig.~\ref{fig:proposed-arch}.
Namely,
given a pair of images, an alignment network estimates the geometric transformation
that aligns them.
The quality of the estimated transformation is assessed using
the proposed \emph{soft-inlier count} which aggregates the observed evidence
in the form of feature matches.
The training objective then is to maximize the alignment quality for
pairs of images which should match.

The key idea is that, instead of requiring strongly supervised training
data in the form of known pairwise alignments and training the alignment
network with these,
the network is ``forced'' into learning to estimate good alignments in order
to achieve high alignment scores (soft-inlier counts) for matching image pairs.
The details of the alignment network and the soft-inlier count are presented next.

\subsection{Semantic alignment network}
\label{sec:alignnet}

In order to make use of the error signal coming from the soft-inlier count,
our framework requires an alignment network which is trainable end-to-end.
We build on the Siamese CNN architecture described in~\cite{Rocco17}, illustrated in the left section of Fig.\,\ref{fig:proposed-arch}.
The architecture is composed of three main stages --
feature extraction, followed by feature matching and
geometric transformation estimation --
which we review below. 

\paragraph{Feature extraction.} 
The input source and target images, $(I^s,I^t)$, are passed through two fully-convolutional feature extraction CNN branches, $F$, with shared weights.
The resulting feature maps $(f^s,f^t)$ are $h\times w\times d$ tensors which can be interpreted as dense $h\times w$ grids of $d$-dimensional local features $f_{ij\mathbf{:}}\in\mathbb{R}^d$.  These individual $d$-dimensional features are L2 normalized.

\paragraph{Pairwise feature matching.}
This stage computes all pairwise similarities, or match scores,
between local features in the two images. This is done with the normalized correlation function, defined as:

\begin{equation}
\begin{split}
S:\mathbb{R}^{h\times w\times d}\times \mathbb{R}^{h\times w\times d}&\to \mathbb{R}^{h\times w\times h \times w}  \\
\label{eq:S}
\end{split}
\end{equation}
\begin{equation}
s_{ijkl}  = 
S(f^s, f^t)_{ijkl} =
\frac{\langle f^s_{ij:}, f^t_{kl:}\rangle}{
\sqrt{\sum_{a,b}\langle f^s_{ab:}, f^t_{kl:}\rangle^2}}, 
\label{eq:s}
\end{equation}
where the numerator in \eqref{eq:s} computes the \emph{raw} pairwise match scores by computing the dot product between features pairs. The denominator performs a normalization operation with the effect of down-weighing ambiguous matches, by penalizing features from one image which have multiple highly-rated matches in the other image.
This is in line with the classical second nearest neighbour test of Lowe \cite{lowe1999object}.
The resulting tensor $s$ contains all normalized match scores between the source and target features.

\begin{figure*}[t]
\centering
\begin{subfigure}[b]{0.3\textwidth}
\centering
\includegraphics[width=0.8\textwidth]{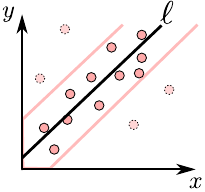}
\caption{Inliers \includegraphics[width=3mm]{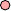} and outliers \includegraphics[width=3mm]{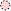} }
\label{fig:ransac-inliers}
\end{subfigure}
\quad
\begin{subfigure}[b]{0.3\textwidth}
\centering
\includegraphics[width=0.8\textwidth]{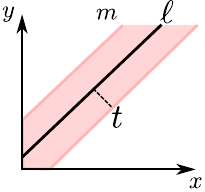}
\caption{Inlier mask function}
\label{fig:ransac-mask}
\end{subfigure}
\quad
\begin{subfigure}[b]{0.3\textwidth}
\centering
\includegraphics[width=0.85\textwidth]{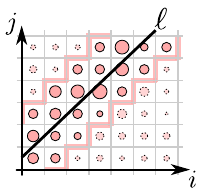}
\caption{Discretized space}
\label{fig:ransac-dense}
\end{subfigure}
\captionsetup{font={small}}
\caption{{\bf Line-fitting example.} (a) The line hypothesis $\ell$ can be evaluated in terms of the number of inliers. (b) The inlier mask $m$ specifies the region where the inlier distance threshold is satisfied.
(c) In the discretized space setting, where the match score $s_{ij}$ exists for every point $(i,j)$, the soft-inlier count is computed by summing up match scores masked by the inlier mask $m$ from (b).}
\label{fig:ransac}
\end{figure*}

\paragraph{Geometric transformation estimation.}
The final stage of the alignment network consists of estimating the parameters of a geometric transformation $g$ given the match scores $s$.
This is done by a transformation regression CNN,
represented by the function $G$:

\begin{equation}
\begin{gathered}
G:\mathbb{R}^{h\times w\times h \times w} \to \mathbb{R}^K, \quad
g = G(s)
\end{gathered}
\label{eq:G}
\end{equation}

\noindent where $K$ is the number of degrees of freedom, or parameters, of the geometric model; \eg $K=6$ for an affine model. The estimated transformation parameters $g$ are used to define the 2-D warping $\mathcal{T}_g$:

\begin{equation}
\begin{gathered}
\mathcal{T}_g: \mathbb{R}^2 \to \mathbb{R}^2, \quad
(u^s,v^s) =  \mathcal{T}_g(u^t,v^t)
\end{gathered}
\end{equation}

\noindent where $(u^t,v^t)$ are the spatial coordinates of the target image, and $(u^s,v^s)$ the corresponding sampling coordinates in the source image. Using $\mathcal{T}_g$, it is possible to warp the source to the target image.

Note that all parts of the geometric alignment network are differentiable
and therefore amenable to end-to-end training~\cite{Rocco17},
including the feature extractor $F$ which can learn better features for the task
of semantic alignment.

\subsection{Soft-inlier count}
\label{sec:softcount}
We propose the \emph{soft-inlier count} used to automatically evaluate the
estimated geometric transformation $g$.
Making an effort to maximize this count provides the weak-supervisory signal required
to train the alignment network, avoiding the need for
expensive manual annotations for $g$.
The soft-inlier count is inspired by the inlier count used in the robust RANSAC method~\cite{fischler1981random}, which is reviewed first.

\paragraph{RANSAC inlier count.} For simplicity, let us consider the problem of fitting a line to a set of observed points $p_i$, with $i=1,\dots N$, as illustrated in Fig.\,\ref{fig:ransac-inliers}.
RANSAC proceeds by sampling random pairs of points used to propose line hypotheses,
each of which is then scored using the inlier count, and the highest scoring line
is chosen; here we only focus on the inlier count aspect of RANSAC used to score
a hypothesis.
Given a hypothesized line $\ell$, the RANSAC inlier scoring function counts the number of observed points which are in agreement with this hypothesis, called the {\em inliers}.
A point $p$ is typically deemed to be an inlier iff its distance to the line is smaller
than a chosen distance threshold $t$, \ie $\text{d}(p,\ell) < t$.

The RANSAC inlier count, $c_R$, can be formulated by means of an auxiliary indicator function illustrated in Fig.\,\ref{fig:ransac-mask}, which we call the inlier mask function $m$:
\begin{equation}
c_R = \sum_{i} m(p_i), \text{ where }
m(p) = 
\begin{cases}
1, &  \text{if } \text{d}(p,\ell)< t \\
0, &  \text{otherwise}.
\end{cases}
\label{eq:Mransac}
\end{equation}

\paragraph{Soft-inlier count.}
The RANSAC inlier count cannot be used directly in a neural network as it is not
differentiable.
Furthermore, in our setting there is no sparse set of matching points,
but rather a match score for every match in a discretized match space.
Therefore, we propose a direct extension, the \emph{soft-inlier count},
which, instead of counting over a sparse set of matches,
sums the match scores over all possible matches.

The running line-fitting example can now be revisited under the
discrete-space conditions, as illustrated in Figure\,\ref{fig:ransac-dense}.
The proposed soft-inlier count for this case is:

\begin{equation}
c = \sum_{i,j} s_{ij}m_{ij},
\end{equation}

\noindent where $s_{ij}$ is the match score at each grid point (i,j), and $m_{ij}$ is the discretized inlier mask:

\begin{equation}
m_{ij}  = \begin{cases} 
1 & \text{if }  \text{d}\big((i,j),\ell\big) < t \\
0       & \text{otherwise}
\end{cases}
\end{equation}

Translating the discrete-space line-fitting example to our semantic alignment problem,
$s$ is a 4-D tensor containing scores for all pairwise feature matches between
the two images (Section~\ref{sec:alignnet}), and
matches are deemed to be inliers if they fit the estimated geometric transformation $g$. 
More formally, the inlier mask $m$ is now also a 4-D tensor, constructed by thresholding
the {\color{black}transfer error}:

\begin{equation}
m_{ijkl}  = \begin{cases} 
1 & \text{if } \text{d}\big((i,j),\mathcal{T}_g(k,l)\big) < t \\
0       & \text{otherwise},
\end{cases}
\end{equation}

\noindent where $\mathcal{T}_g(k,l)$ are the estimated coordinates of
target image's point $(k,l)$ in the source image according to the
geometric transformation $g$;
$\text{d}\big((i,j),\mathcal{T}_g(k,l)\big)$ is the {transfer error}
as it measures how aligned is the point $(i,j)$ in the source image,
with the projection of the target image point $(k,l)$ into the source image.
The soft-inlier count $c$ is then computed by summing the masked matching scores over the entire space of matches:
\begin{equation}
c = \sum_{i,j,k,l} s_{ijkl}m_{ijkl}.
\end{equation}

\paragraph{Differentiability.} The proposed soft-inlier count $c$ is differentiable with respect to the transformation parameters $g$  as long as the geometric transformation $\mathcal{T}_g$ is differentiable~\cite{Jaderberg15}, which is the case for a range of standard geometric transformations such as 2D affine, homography or thin-plate spline transformations.
Furthermore, it is also differentiable w.r.t. the match scores, which facilitates
training of the feature extractor.

\paragraph{Implementation as a CNN layer.}
The inlier mask $m$ can be computed by warping an identity mask $m^{\mathit{Id}}$ with the estimated transformation $\mathcal{T}_g$, where $m^{\mathit{Id}}$ is constructed by thresholding the {transfer error} of the identity transformation:

\begin{equation}
m^{\mathit{Id}}_{ijkl}=
\begin{cases}
1 & \text{d}\big((i,j),(k,l)\big) < t \\
0 & \text{otherwise}.
\end{cases}
\end{equation}

The warping is implemented using a spatial transformer layer~\cite{Jaderberg15}, which consists of a  grid generation layer and a bilinear sampling layer. Both of these functions are readily available in most deep learning frameworks.

\paragraph{Optimization objective.}
For a given training pair of images that should match, the goal is to maximize
the soft-inlier count $c$, or, equivalently, to minimize the loss $\mathcal{L}=-c$.

\paragraph{Analogy to RANSAC.}
Please also note that our method is similar in spirit to RANSAC~\cite{fischler1981random}, where (i) transformations are proposed (by random sampling) and then (ii) scored by their support (number of inliers). In our case, during training (i) the transformations are proposed (estimated) by the regressor network $G$ and (ii) scored using the proposed soft-inlier score.  The gradient of this score is used to improve both the regressor $G$ and feature extractor $F$ (see Fig.\,\ref{fig:proposed-arch}).  In turn, the regressor produces better transformations  and the feature extractor better feature matches that maximize the soft-inlier score on training images.

\section{Evaluation and results}
In this section we provide implementation details,
benchmarks used to evaluate our approach, and quantitative and qualitative results.

\subsection{Implementation details}
\label{sec:impl}

\paragraph{Semantic alignment network.}
For the underlying semantic alignment network,
we use the best-performing architecture from~\cite{cnngeometric} which employs a ResNet-101~\cite{he2016deep}, cropped after \texttt{conv4-23}, as the feature extraction CNN $F$.
Note that this is a better performing model than the one described in~\cite{Rocco17}, mainly due to use of ResNet versus VGG-16~\cite{Simonyan15}.
Given an image pair, the model produces a thin-plate spline geometric transformation $\mathcal{T}_g$ which aligns the two images;
$\mathcal{T}_g$ has 18 degrees of freedom.
The network is initialized with the pre-trained weights from~\cite{cnngeometric},
and we finetune it with our weakly supervised method.
{
Note that the initial model has been trained in a self-supervised way from synthetic data, not requiring human supervision \cite{Rocco17},
therefore not affecting our claim of weakly supervised training\footnote{The initial model is trained with a supervised loss, but the ``supervision'' is automatic due to the use
of synthetic data.}.}

\begin{table*}[t!]
\footnotesize

\hspace{-1mm}\begin{tabular}{@{\hskip 1mm}l@{\hskip 1.4mm}c@{\hskip 1.4mm}c@{\hskip 1.4mm}c@{\hskip 1.4mm}c@{\hskip 1.4mm}c@{\hskip 1.4mm}c@{\hskip 1.4mm}c@{\hskip 1.4mm}c@{\hskip 1.4mm}c@{\hskip 1.4mm}c@{\hskip 1.4mm}c@{\hskip 1.4mm}c@{\hskip 1.4mm}c@{\hskip 1.4mm}c@{\hskip 1.4mm}c@{\hskip 1.4mm}c@{\hskip 1.4mm}c@{\hskip 1.4mm}c@{\hskip 1.4mm}c@{\hskip 1.4mm}c@{\hskip 1.4mm} | c @{\hskip 1mm}}

\noalign{\hrule height 0.7pt}
    \bigstrut[tb]
Method &	aero &	bike &	bird &	boat &	bottle &	bus &	car &	cat &	chair &	cow &	d.table &	dog &	horse &	moto &	person &	plant &	sheep &	sofa &	train &	tv & 	{\bf all}	\\			
\hhline{=====================|=}
  \bigstrut[t]
HOG+PF-LOM~\cite{ham2016} &	73.3 &	74.4 &	54.4 &	50.9 &	49.6 &	73.8 &	72.9 &	63.6 &	46.1 &	79.8 &	42.5 &	48.0 &	68.3 &	66.3 &	42.1 &	62.1 &	65.2 &	57.1 &	64.4 &	58.0 &	62.5	\\
VGG-16+SCNet-A~\cite{han2017scnet} &	67.6 &	72.9 &	69.3 &	59.7 &	{\bf 74.5} &	72.7 &	73.2 &	59.5 &	51.4 &	78.2 &	39.4 &	50.1 &	67.0 &	62.1 &	69.3 &	68.5 &	78.2 &	63.3 &	57.7 &	59.8 &	66.3	\\
VGG-16+SCNet-AG~\cite{han2017scnet} &	83.9 &	81.4 &	70.6 &	62.5 &	60.6 &	81.3 &	81.2 &	59.5 &	53.1 &	81.2 &	{\bf 62.0} &	58.7 &	65.5 &	73.3 &	51.2 &	58.3 &	60.0 &	69.3 &	61.5 &	{\bf 80.0} &	69.7	\\
VGG-16+SCNet-AG+~\cite{han2017scnet} &	{\bf 85.5} &	84.4 &	66.3 &	{\bf 70.8} &	57.4 &	82.7 &	82.3 &	71.6 &	{\bf 54.3} &	{\bf 95.8} &	55.2 &	59.5 &	68.6 &	75.0 &	56.3 &	60.4 &	60.0 &	{\bf 73.7} &	{\bf 66.5} &	76.7 &	72.2	\\
VGG-16+CNNGeo~\cite{Rocco17} &	75.2 &	80.1 &	73.4 &	59.7 &	43.8 &	77.9 &	84.0 &	67.7 &	44.3 &	89.6 &	33.9 &	67.1 &	60.5 &	72.6 &	54.0 &	41.0 &	60.0 &	45.1 &	58.3 &	37.2 &	65.0	\\
ResNet-101+CNNGeo~\cite{Rocco17} &	82.4 &	80.9 &	{\bf 85.9} &	47.2 &	57.8 &	83.1 &	{\bf 92.8} &	{\bf 86.9} &	43.8 &	91.7 &	28.1 &	{\bf 76.4} &	70.2 &	{\bf 76.6} &	68.9 &	65.7 &	{\bf 80.0} &	50.1 &	46.3 &	60.6 &	71.9	\\
Proposed &	83.7 &	{\bf 88.0} &	83.4 &	58.3 &	68.8 &	{\bf 90.3} &	92.3 &	83.7 &	47.4 &	91.7 &	28.1 &	76.3 &	{\bf 77.0} &	76.0 &	{\bf 71.4} &	{\bf 76.2} &	{\bf 80.0} &	59.5 &	62.3 &	63.9 &	{\bf 75.8}	\\

\noalign{\hrule height 0.7pt}

\end{tabular}
\captionsetup{font={small}}
\vspace{-1mm}
\caption{{\bf Per-class PCK on the PF-PASCAL dataset.}}
\label{tab:eval_pf_pascal}
\vspace{-1mm}
\end{table*}																									

\paragraph{Training details.}
Training and validation image pairs are obtained from the training set of PF-PASCAL, described in Section~\ref{sec:eval_benchmarks}. 
All input images are resized to $240 \times 240$, {and the value $t=L/30$ (where $L=h=w$ is the size of the extracted feature maps) was used for the {transfer error} threshold.}
The whole model is trained end-to-end, including the affine parameters in the batch normalization layers. However, the running averages of the batch normalization layers are kept fixed, in order to be less dependent on the particular statistics of the training dataset.
The network is implemented in PyTorch~\cite{pytorch}
and trained using
the Adam optimizer~\cite{kingma2015adam} with learning rate $5\cdot10^{-8}$, no weight decay and batch size of 16. 
The training dataset is augmented by horizontal flipping, swapping the source and target images, and random cropping. 
Early stopping is {required} to avoid overfitting, {given the small size of the training set. This results in
13 training epochs, taking about an hour on a modern GPU.}

\subsection{Evaluation benchmarks\label{sec:eval_benchmarks}}
Evaluation is performed on three standard image alignment benchmarks: PF-PASCAL, Caltech-101 and TSS.

\paragraph{PF-PASCAL~\cite{ham2017}.} This dataset contains 1351 semantically related image pairs from 20 object categories, which present challenging appearance differences and background clutter. We use the split proposed in~\cite{han2017scnet}, which divides the dataset into roughly 700 pairs for training, 300 pairs for validation, and 300 pairs for testing. Keypoint annotations are provided for each image pair, which are used only for evaluation purposes.
Alignment quality is evaluated in terms of the percentage of correct keypoints (PCK) metric~\cite{Yang13}, which counts the number of keypoints which have a {transfer error} below a given threshold. 
{We follow the procedure employed in~\cite{han2017scnet}, where keypoint $(x,y)$ coordinates are normalized in the $[0,1]$ range by dividing with the image width and height respectively, and the value $\alpha=0.1$ is employed as the distance threshold.}

\paragraph{Caltech-101~\cite{fei2006one}.} Although originally introduced for the image classification task, the dataset was adopted in~\cite{Kim13} for assessing semantic alignment, and has been then extensively used for this purpose~\cite{ham2017,kim2017fcss,han2017scnet,Rocco17}. The evaluation is performed on 1515 semantically related image pairs, 15 pairs for each of the 101 object categories of the dataset.
The semantic alignment is evaluated using three different metrics: (i) the label transfer accuracy (LT-ACC); (ii) the intersection-over-union (IoU), and; (iii) the object localization error (LOC-ERR). The label transfer accuracy and the intersection-over-union both measure the overlap between the annotated foreground object segmentation masks, with former putting more emphasis on the background class and the latter on the foreground object. The localization error computes a dense displacement error. However, given the lack of dense displacement annotations, the metric computes the ground-truth transformation
from the source and target bounding boxes, thus assuming that the transformation
is a simple translation with axis-aligned anisotropic scaling.
This assumption is unrealistic as, amongst others, it does not cover rotations, affine or deformable transformations.
Therefore, we believe that LOC-ERR should not be reported any more, but report it here for completeness and in order to adhere to the currently adopted evaluation protocol.

\paragraph{TSS~\cite{taniai2016joint}.}
The recently introduced TSS dataset contains 400 semantically related image pairs, which are split into three different subsets: FG3DCar, JODS and PASCAL, according to the origin of the images. Ground-truth flow is provided for each pair, which was obtained by manual annotation of sparse keypoints, followed by automatic densification using an interpolation algorithm. The evaluation metric is the PCK computed densely over the foreground object. The distance threshold is defined as $\alpha\,\text{max}(w^s,h^s)$ with $(w^s,h^s)$ being the dimensions of the source image, and $\alpha=0.05$.

\paragraph{Assessing generalization.}
We train a single semantic alignment network with the 700 training pairs from PF-PASCAL \emph{without} using the keypoint annotations,
and stress that our weakly-supervised training objective only uses
the information that the image pair should match.
\emph{The same} model is then used for all experiments -- evaluation on the
test sets of PF-PASCAL, Caltech-101 and TSS datasets.
This poses an additional difficulty as these datasets contain images of different object categories or of different nature. While PF-PASCAL contains images of common objects such as car, bicycle, boat, \etc, Caltech-101 contains images of much less common categories such as accordion, buddha or octopus. On the other hand, while the classes of TSS do appear in PF-PASCAL,
the pose differences in TSS are usually smaller than in PF-PASCAL, which modifies the challenge into obtaining a very precise alignment.

\subsection{Results}

In the following, our alignment network trained with \emph{weak supervision}
is compared to the state-of-the-art alignment methods, many of which require
\emph{manual annotations} or \emph{strong supervision}
(\cf Table~\ref{tab:rel_work}).

\paragraph{PF-PASCAL.} From Table \ref{tab:eval_pf_pascal} it is clear
that our method sets the new state-of-the-art, achieving an overall PCK of 
75.8\%,
which is a
3.6\%
improvement over the best competitor \cite{han2017scnet}.
This result is impressive as the two methods are trained on the same image pairs,
with ours being weakly supervised while \cite{han2017scnet} make use of
bounding box annotations.

The benefits of weakly supervised training can be seen by comparing our method
with ResNet-101+CNNGeo \cite{Rocco17,cnngeometric}.
The two use the same base alignment network (\cf Section~\ref{sec:impl}),
but ResNet-101+CNNGeo was trained only on synthetically deformed image pairs,
while ours employs the proposed weakly supervised fine-tuning.
The 
3.9\% boost clearly demonstrates the advantage obtained by training
on real image pairs and thus encountering rich appearance variations,
as opposed to using synthetically transformed pairs in ResNet-101+CNNGeo \cite{Rocco17}.

\begin{table}[t!]
\small
\begin{center}
\begin{tabular}{lccc}
\toprule				
Method &	LT-ACC &	IoU &	LOC-ERR	\\ \midrule \midrule
HOG+PF-LOM~\cite{ham2017} &	0.78 &	0.50 &	0.26 \\
FCSS+SIFT Flow~\cite{kim2017fcss} &	0.80 &	0.50 &	0.21 \\
FCSS+PF-LOM~\cite{kim2017fcss} &	0.83 &	0.52 &	0.22 \\
VGG-16+SCNet-A~\cite{han2017scnet} &	0.78 &	0.50 &	0.28 \\
VGG-16+SCNet-AG~\cite{han2017scnet} &	0.78 &	0.50 &	0.27 \\
VGG-16+SCNet-AG+~\cite{han2017scnet} &	0.79 &	0.51 &	0.25 \\
HOG+OADSC~\cite{yang2017object} &	0.81 &	0.55 &	{\bf 0.19}	\\
VGG-16+CNNGeo~\cite{Rocco17} &	0.80 &	0.55 &	0.26 \\
ResNet-101+CNNGeo~\cite{Rocco17} &	0.83 &	0.61 &	0.25 \\
Proposed &	{\bf 0.85} &	{\bf 0.63} &	0.24 \\
\bottomrule
\end{tabular}
\captionsetup{font={small}}
\vspace{-1mm}
\captionof{table}{{\bf Evaluation results on the Caltech-101 dataset.}}
\label{tab:eval_caltech}
\vspace{-3mm}
\end{center}
\end{table}				

\paragraph{Caltech-101.}
Table \ref{tab:eval_caltech} presents the quantitative results for this dataset.
The proposed method beats state-of-the-art results in terms of the label-transfer accuracy and intersection-over-union metrics.
Weakly supervised training again improves the results, by 2\%, over the synthetically trained ResNet-101+CNNGeo.
In terms of the localization-error metric, our model does not attain state-of-the-art performance,
but we argue that this metric is not a good indication of the alignment quality, as explained in section \ref{sec:eval_benchmarks}.
This claim is further backed up by noticing that the relative ordering of various methods
based on this metric is in direct opposition with the other two metrics.

\paragraph{TSS.} The quantitative results for the TSS dataset are presented in Table \ref{tab:eval_tss}.
We set the state-of-the-art for two of the three subsets of the TSS dataset: FG3DCar and JODS.
Although our weakly supervised training provides an improvement over the base alignment network, ResNet-101+CNNGeo, the gain is modest.
We believe the reason is a very different balancing of classes in this dataset compared to our training. Recall our model is trained \emph{only once} on the PF-PASCAL dataset,
and is then applied without any further training on TSS and Caltech-101.

\begin{figure}[t!]		
\centering
\begin{center}		
\setlength{\tabcolsep}{1pt} 
\renewcommand{\arraystretch}{1} 
 \begin{subfigure}[t]{\columnwidth}
 \centering
\begin{tabular}{@{\hskip 0mm}c}		
\resizebox{0.97\columnwidth}{!}{\includegraphics[height=2.5cm]{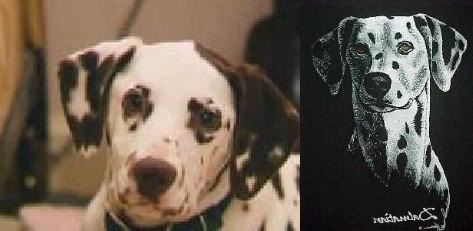} 	\includegraphics[height=2.5cm]{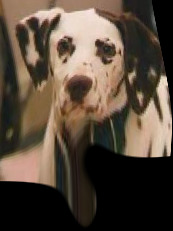}} \\
\resizebox{0.97\columnwidth}{!}{\includegraphics[height=1.8cm]{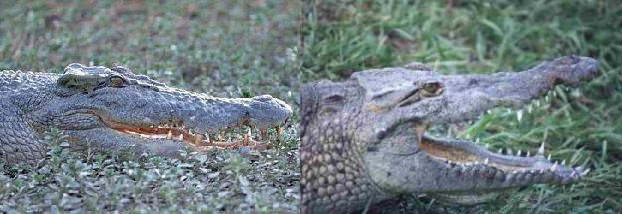} 	\includegraphics[height=1.8cm]{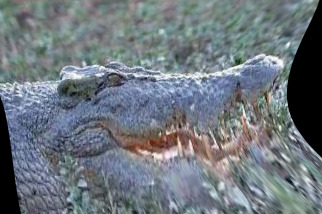}} \\
\addlinespace[-0.5ex]
 \end{tabular}
 \vspace{-1.5mm}
 \caption{Caltech-101 \label{fig:qual_caltech}} 
\end{subfigure}

  \vspace{2mm}
  
\begin{subfigure}[t]{\columnwidth}
\centering
\begin{tabular}{@{\hskip 0mm}c}	
\resizebox{0.97\columnwidth}{!}{\includegraphics[height=2cm]{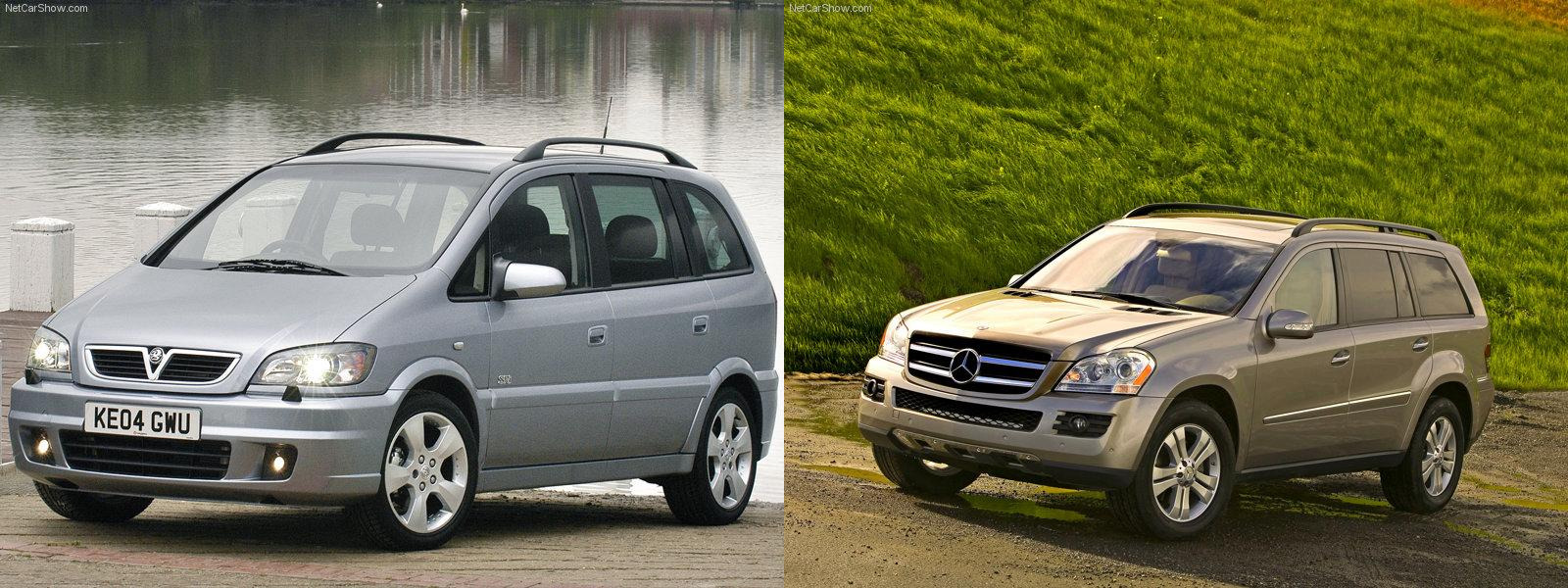} 	\includegraphics[height=2cm]{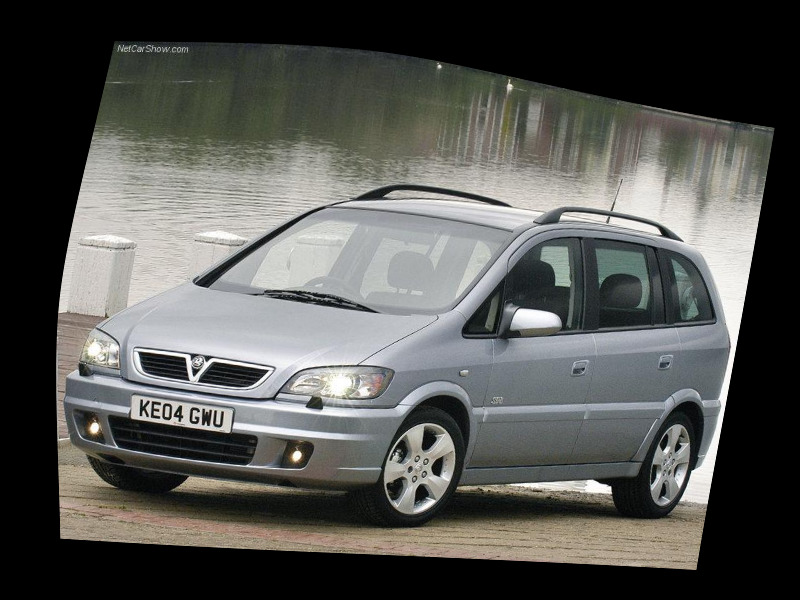}} \\
\resizebox{0.97\columnwidth}{!}{\includegraphics[height=2cm]{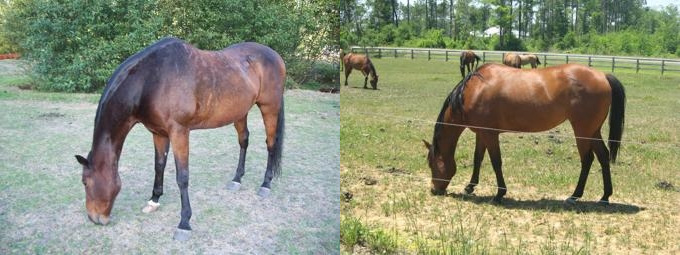} 	\includegraphics[height=2cm]{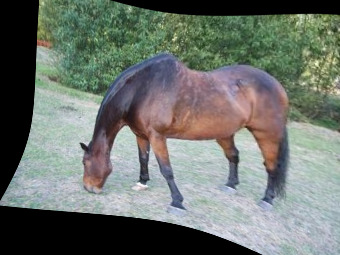}} \\
\resizebox{0.97\columnwidth}{!}{\includegraphics[height=1.9cm]{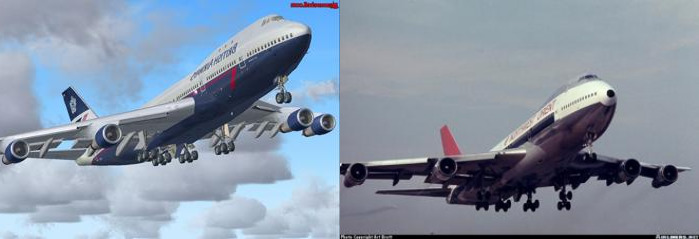} 	\includegraphics[height=1.9cm]{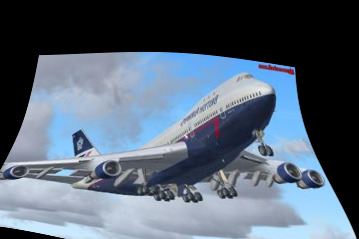}} \\
\resizebox{0.97\columnwidth}{!}{\includegraphics[height=2cm]{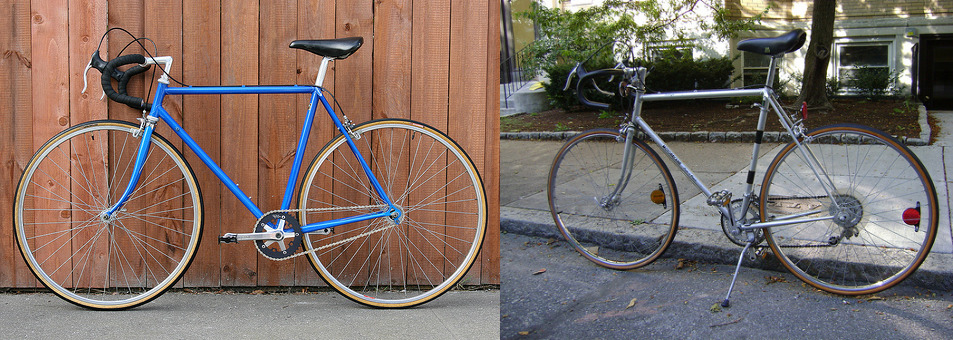} 	\includegraphics[height=2cm]{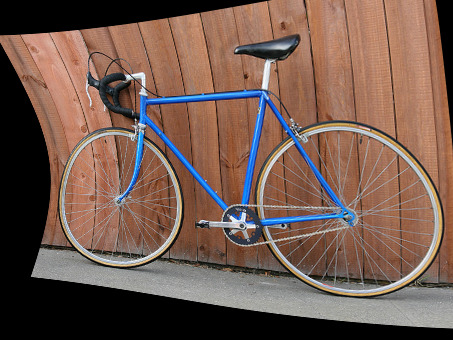}} \\
\end{tabular}	
 \vspace{-1.5mm}
 \caption{TSS \label{fig:qual_tss}} 
\end{subfigure}
\captionsetup{font={small}}	
\vspace{-2mm}	
\caption{{\bf Alignment examples on the Caltech-101 and TSS datasets.} Each row shows the (left) source and (middle) target images, and (right) the automatic semantic alignment. 
}	
\vspace{-7mm}		
\end{center}		
\end{figure}

\begin{figure*}[t!]		
\begin{center}		
\setlength{\tabcolsep}{1pt} 
\renewcommand{\arraystretch}{1} 
\newcommand{\size}{2.5}	
\begin{tabular}{cc}		

\resizebox{0.45\textwidth}{!}{\includegraphics[height=3.2cm]{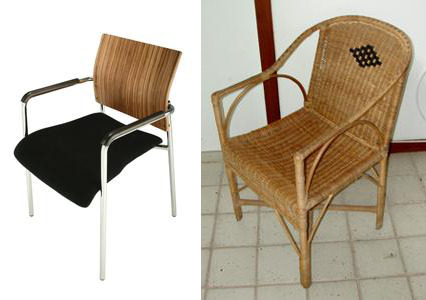} 	\includegraphics[height=3.2 cm]{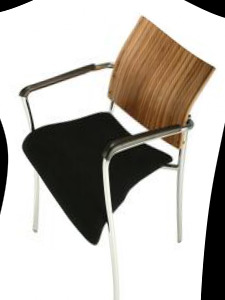}} &	\includegraphics[height=3.8 cm]{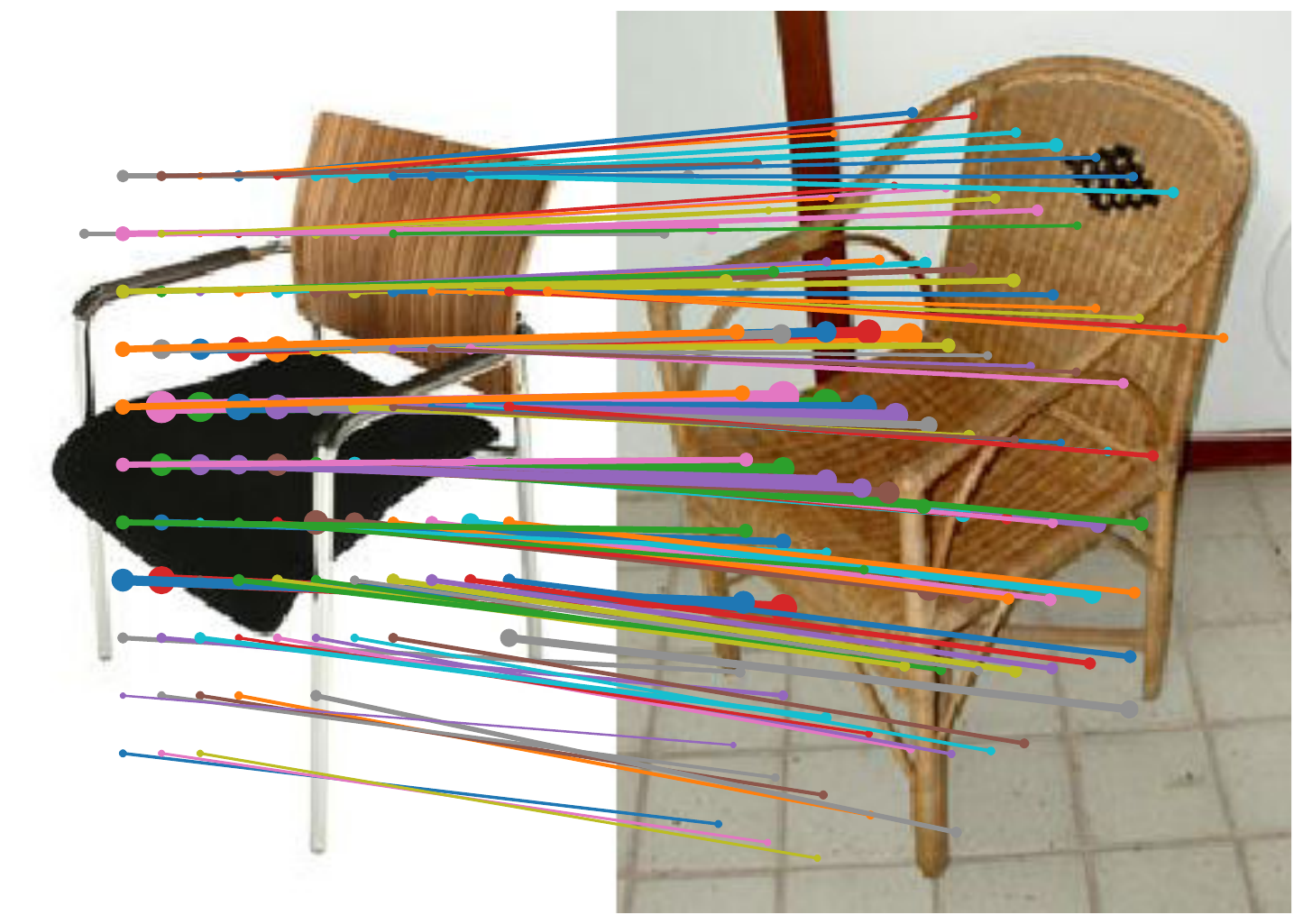} \\
\resizebox{0.45\textwidth}{!}{\includegraphics[height=3.2 cm]{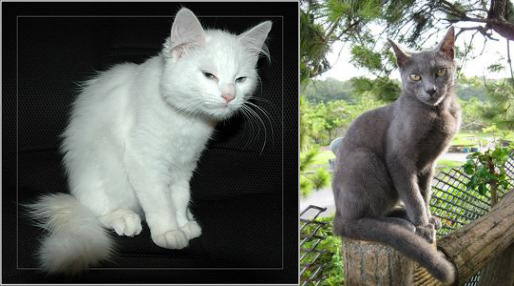} 	\includegraphics[height=3.2 cm]{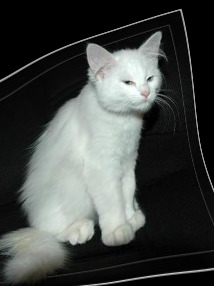}} &	\includegraphics[height=3.1 cm]{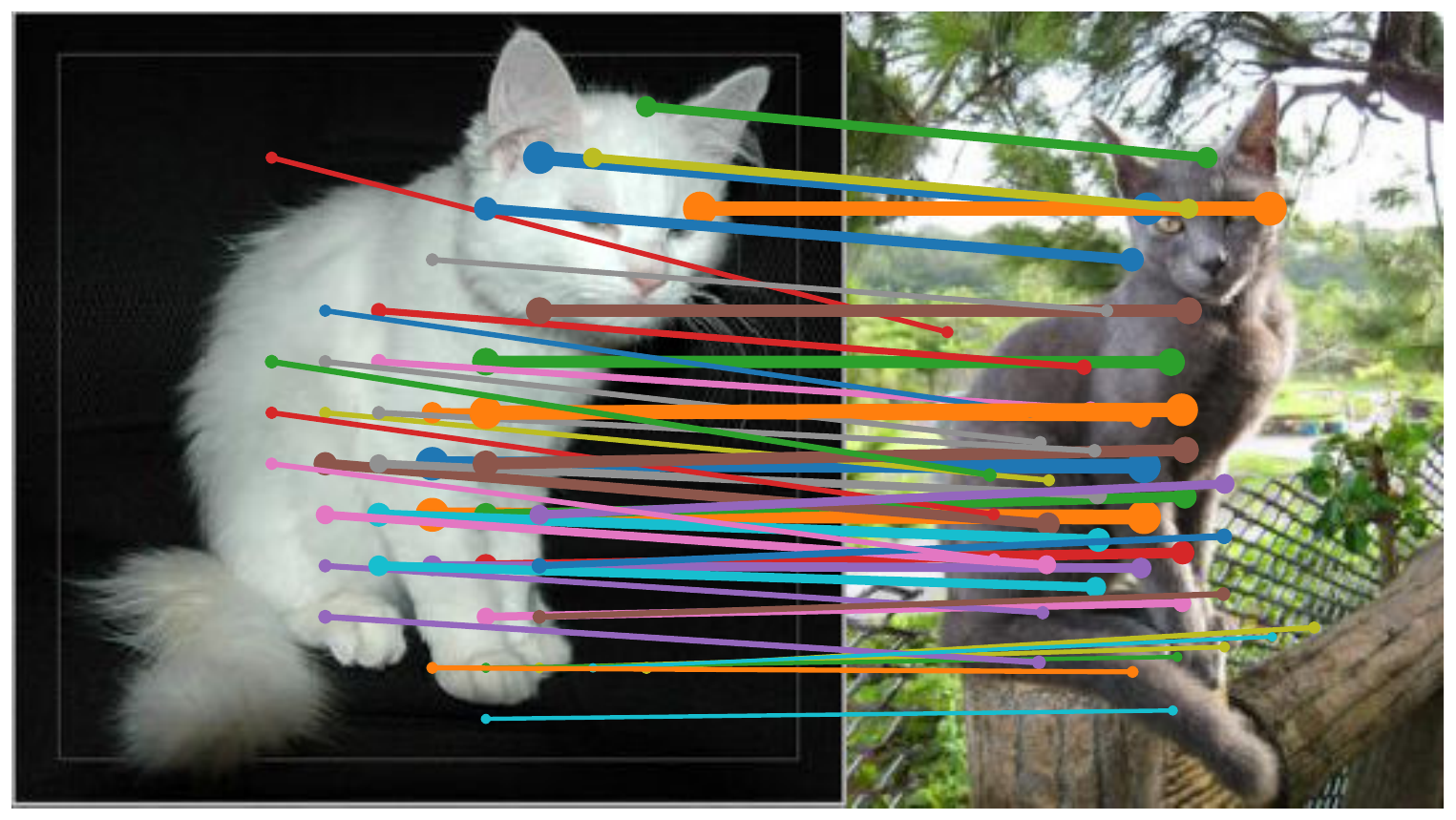} \\

\resizebox{0.57\textwidth}{!}{\includegraphics[height=2.9 cm]{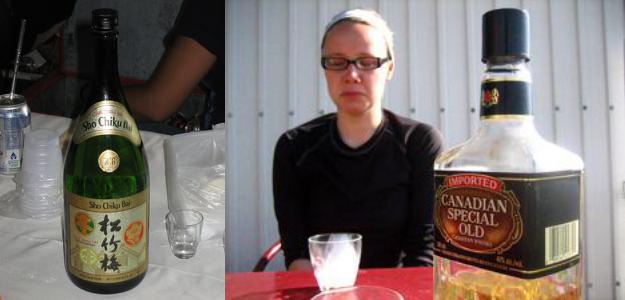} 	\includegraphics[height=2.9 cm]{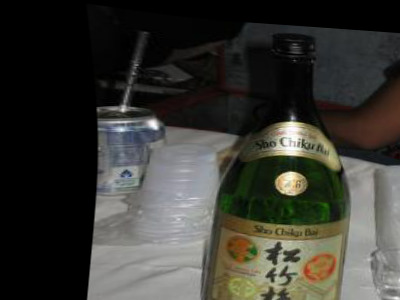}} &	\includegraphics[height=2.9cm]{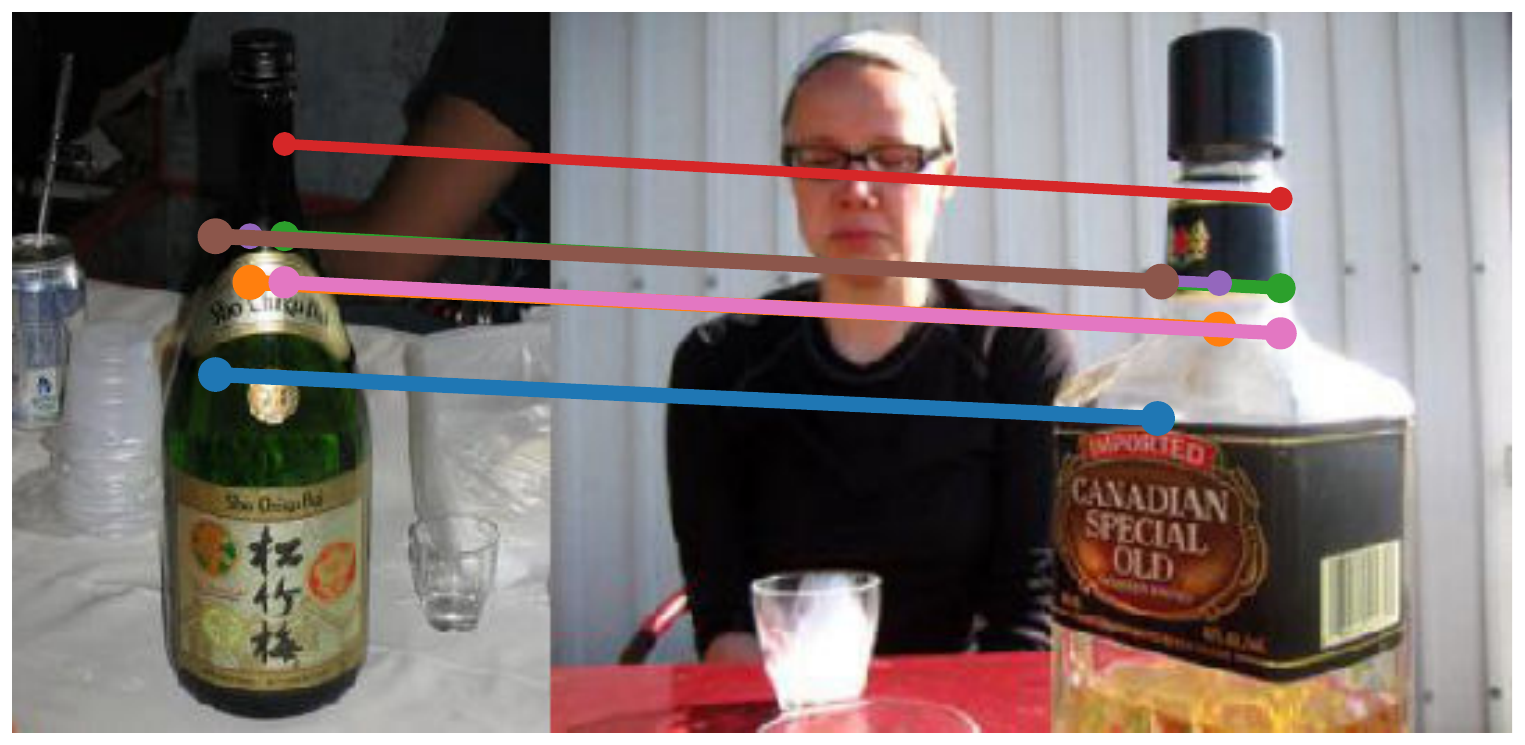} \\
\resizebox{0.57\textwidth}{!}{\includegraphics[height=2.5 cm]{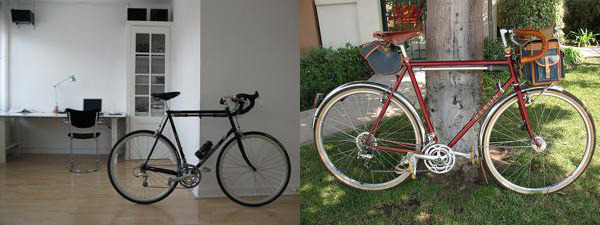} \includegraphics[height=2.5 cm]{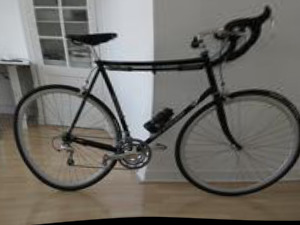}} &	\includegraphics[height=2.5 cm]{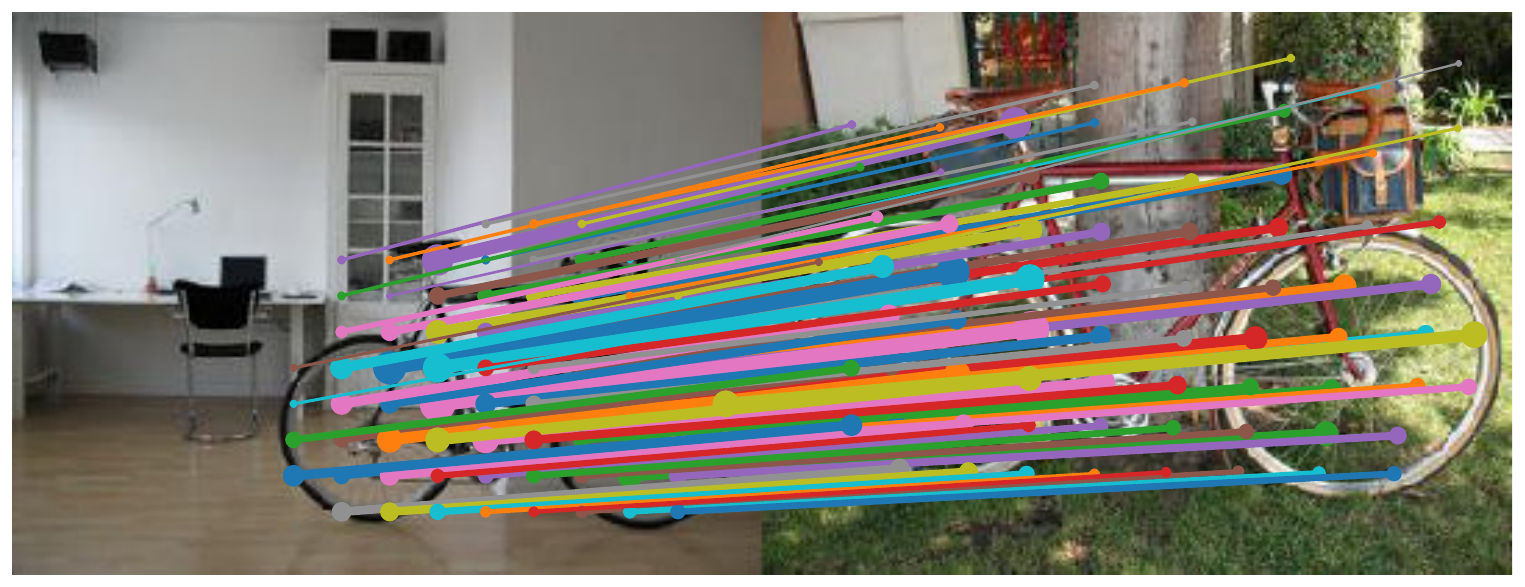} \\
\resizebox{0.57\textwidth}{!}{\includegraphics[height=2.5 cm]{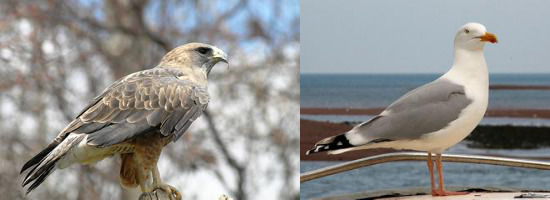} 	\includegraphics[height=2.5 cm]{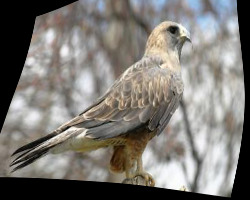}} &	\includegraphics[height=2.5 cm]{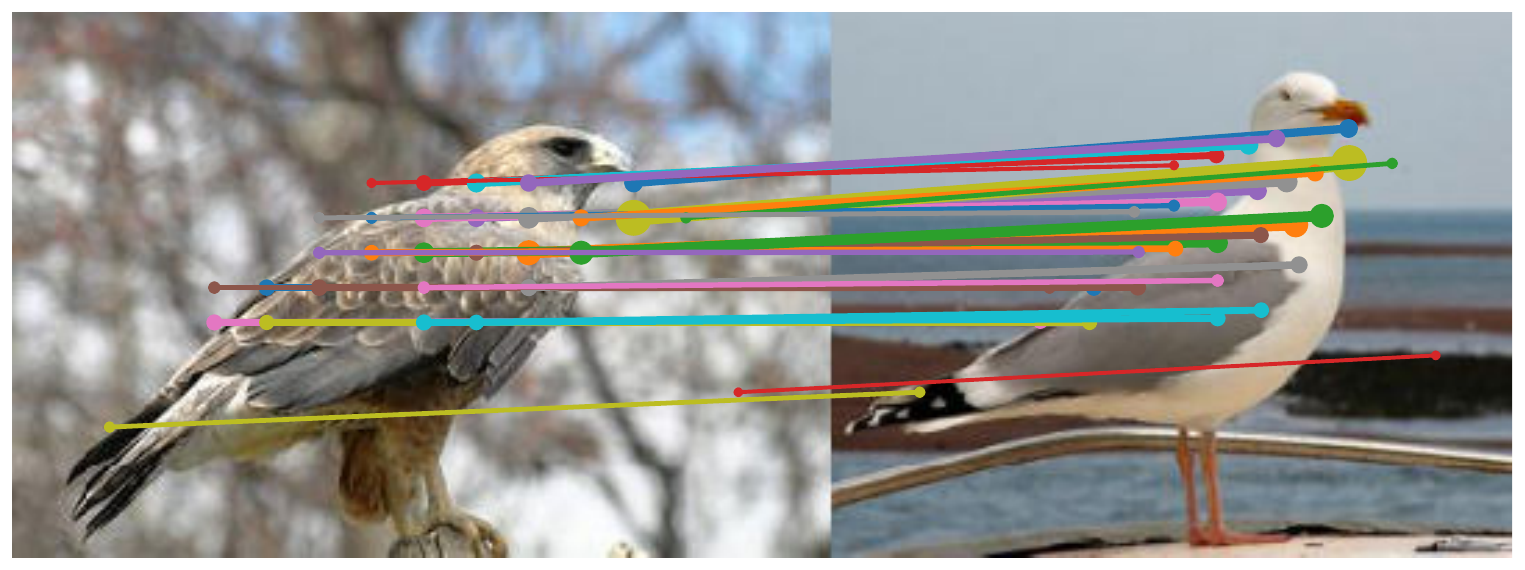} \\
\resizebox{0.57\textwidth}{!}{\includegraphics[height=2.4 cm]{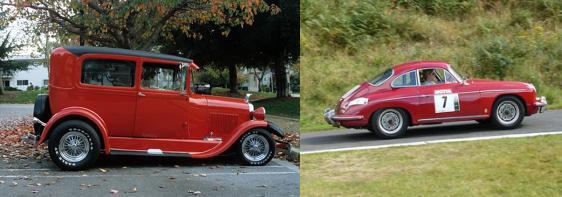} 	\includegraphics[height=2.4 cm]{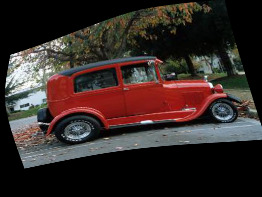}} &	\includegraphics[height=2.4 cm]{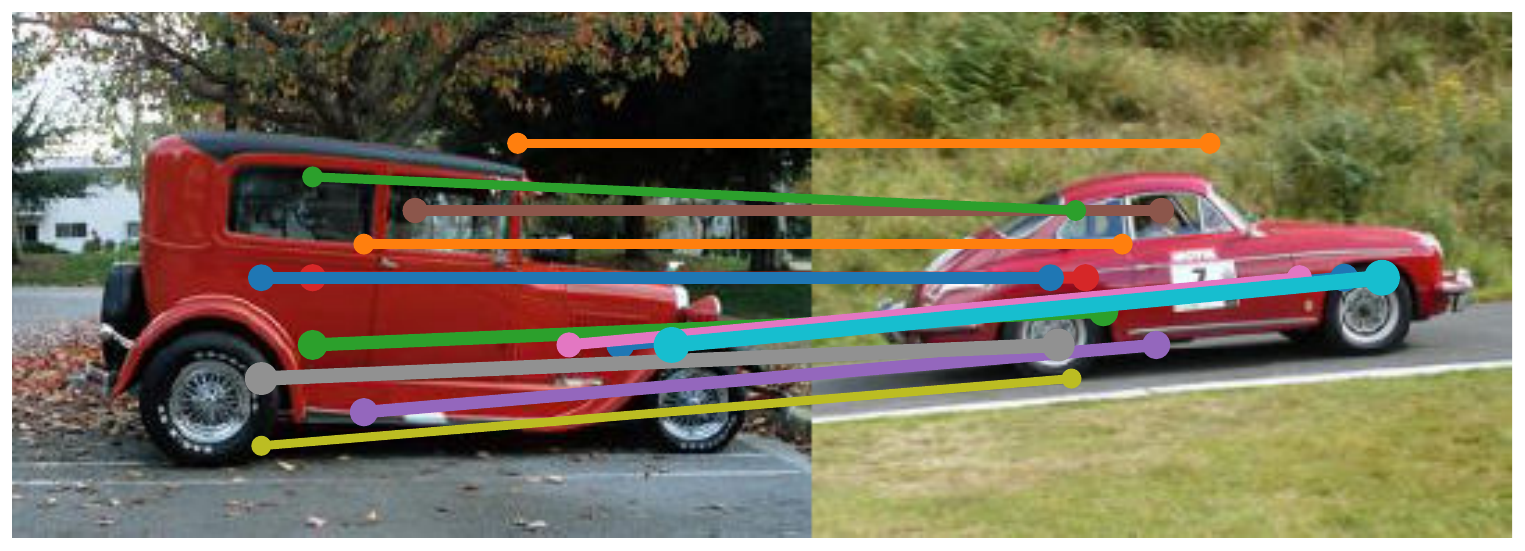} \\
\resizebox{0.57\textwidth}{!}{\includegraphics[height=2.6 cm]{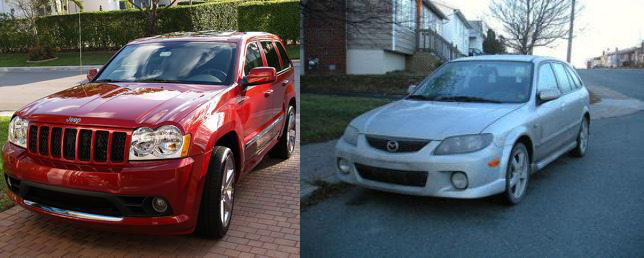} 	\includegraphics[height=2.6 cm]{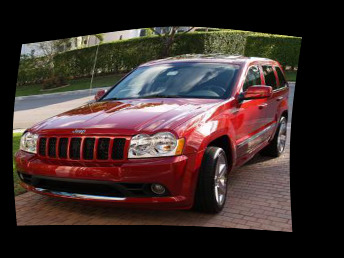}} &	\includegraphics[height=2.6 cm]{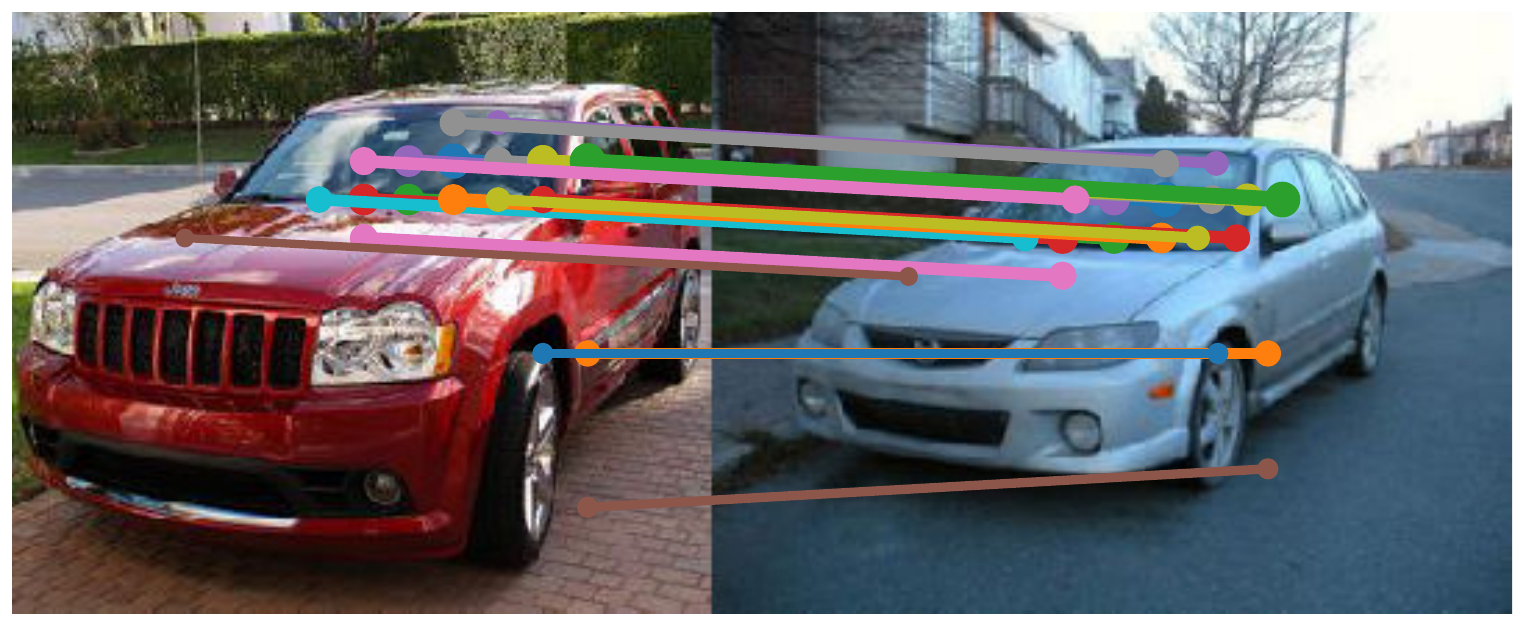} \\
(a) Semantic alignment & (b) Strongest inlier matches \\

\end{tabular}		
\captionsetup{font={small}}		
\caption{{\bf Alignment examples on the PF-PASCAL dataset.} Each row corresponds to one example. (a) shows the (right) automatic semantic alignment of the (left) source and (middle) target images. (b) shows the strongest inlier feature matches.}
\label{fig:qual_pf_pascal}		
\end{center}		
\end{figure*}

\begin{figure*}[t!]		
\begin{center}		
\setlength{\tabcolsep}{1pt} 
\renewcommand{\arraystretch}{1} 
\newcommand{\size}{2.5}	
\begin{tabular}{cc}		

\isArXiv{\includegraphics[height=3.1 cm]{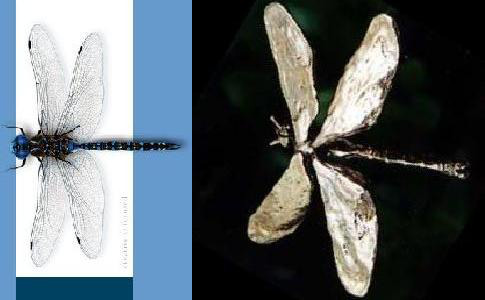}}{
\begin{tikzpicture}\node (label) at (0,0)[anchor=south west,inner sep=0,outer sep=0]{
\includegraphics[height=3.1 cm]{selected-caltech/513_pair}};\node [outer sep=1,inner sep=2,fill=white,opacity=0.7,text opacity=1] (A) at (0.4,0.4) {\Large 8};\end{tikzpicture}}
\includegraphics[height=3.1 cm]{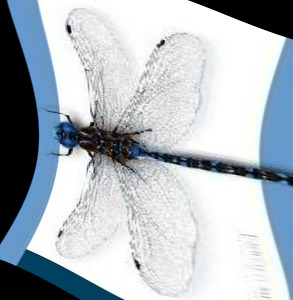} &	\includegraphics[height=3.1 cm]{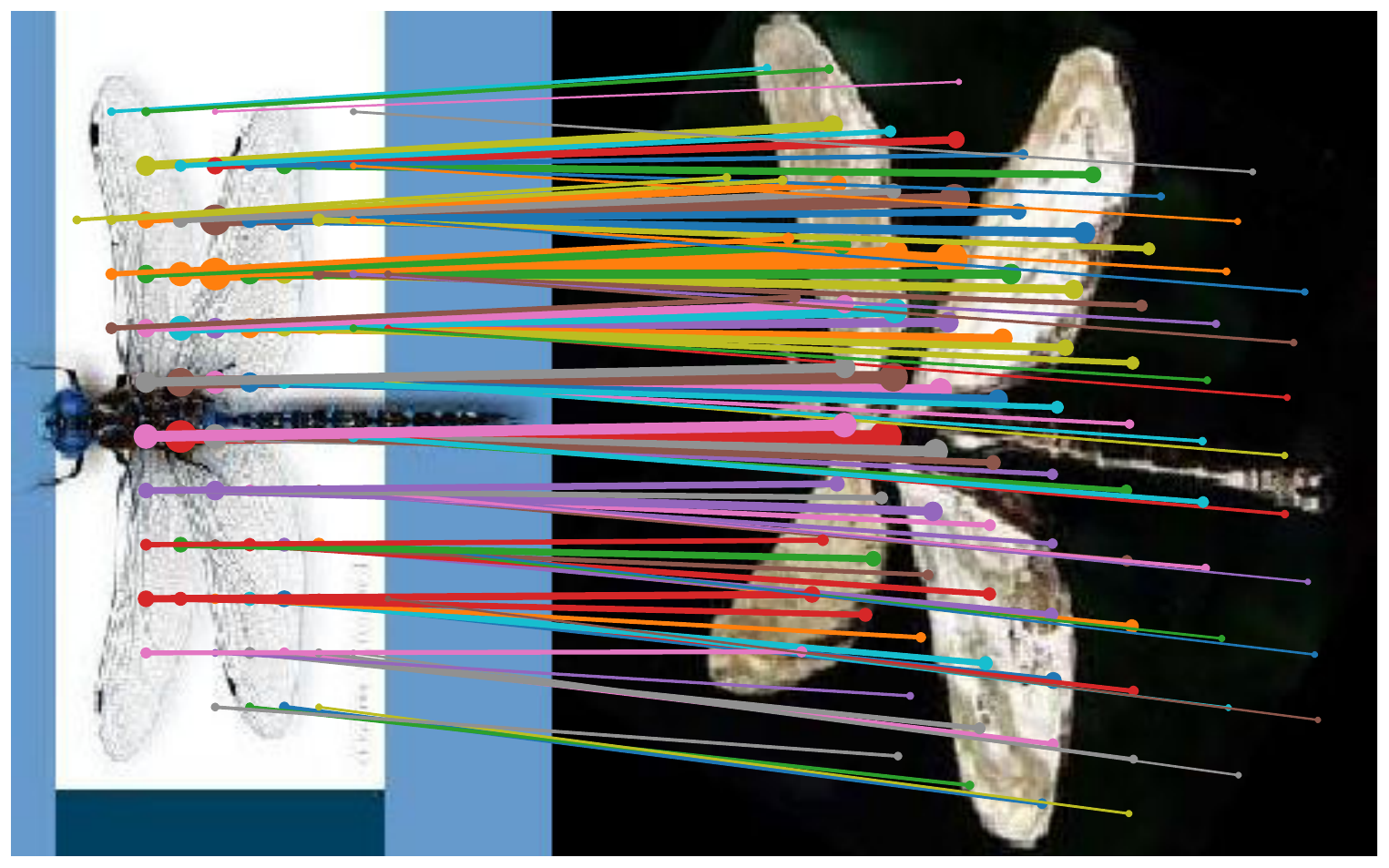} \\

\isArXiv{\includegraphics[height=2.7 cm]{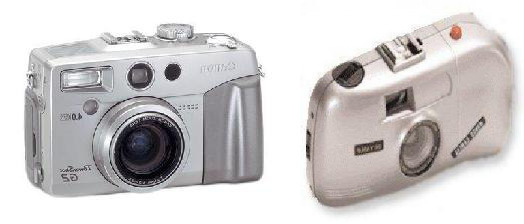}}{
\begin{tikzpicture}\node (label) at (0,0)[anchor=south west,inner sep=0,outer sep=0]{
\includegraphics[height=2.7 cm]{selected-caltech/259_pair}};\node [outer sep=1,inner sep=2,fill=white,opacity=0.7,text opacity=1] (A) at (0.4,0.4) {\Large 9};\end{tikzpicture}}
\includegraphics[height=2.7 cm]{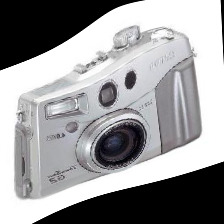} &	\includegraphics[height=2.7 cm]{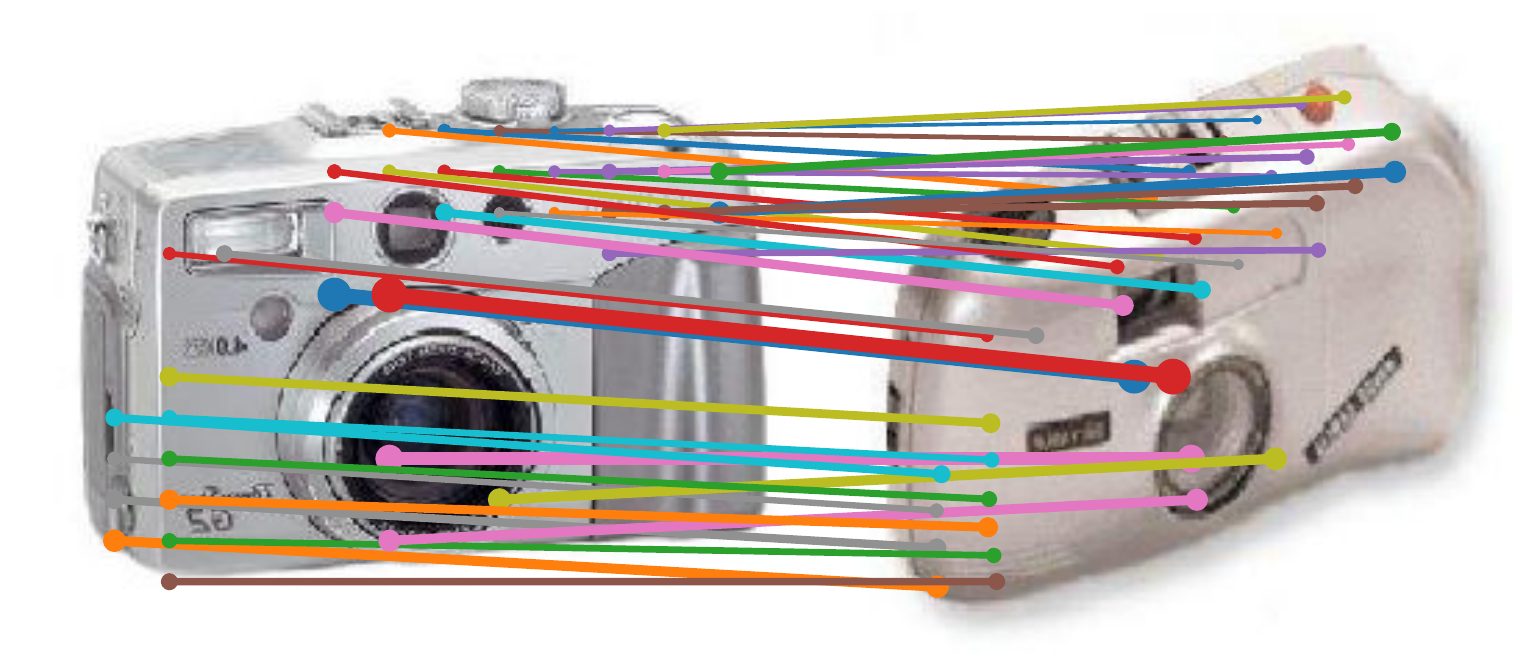} \\

\isArXiv{\includegraphics[height=3.1 cm]{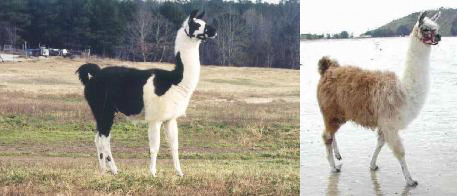}}{
\begin{tikzpicture}\node (label) at (0,0)[anchor=south west,inner sep=0,outer sep=0]{
\includegraphics[height=3.1 cm]{selected-caltech/875_pair}};\node [outer sep=1,inner sep=2,fill=white,opacity=0.7,text opacity=1] (A) at (0.4,0.4) {\Large 10};\end{tikzpicture}}
\includegraphics[height=3.1 cm]{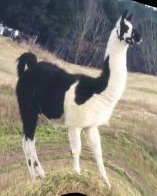} &	\includegraphics[height=3.1 cm]{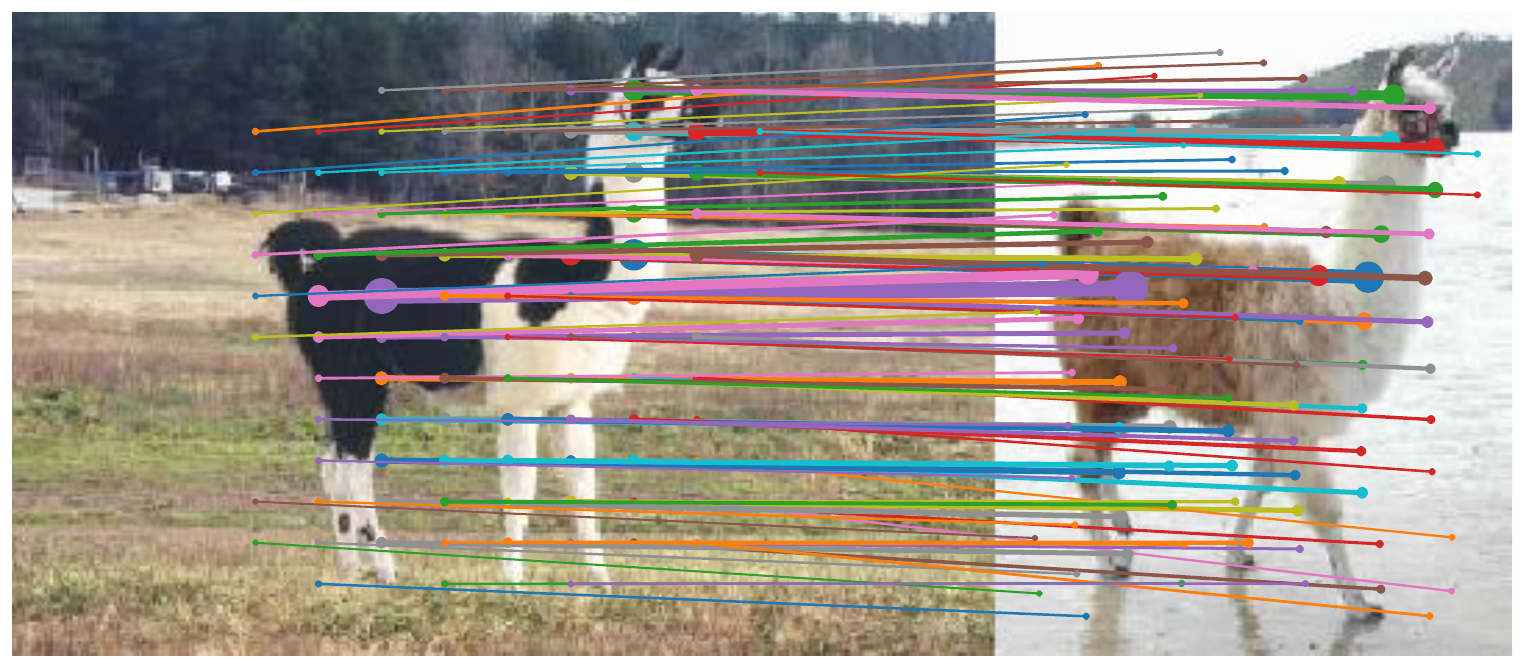} \\

\isArXiv{\includegraphics[height=3.1 cm]{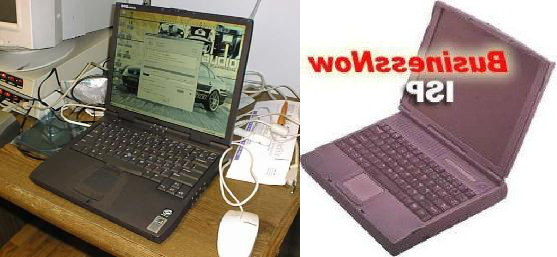}}{
\begin{tikzpicture}\node (label) at (0,0)[anchor=south west,inner sep=0,outer sep=0]{
\includegraphics[height=3.1 cm]{selected-caltech/860_pair}};\node [outer sep=1,inner sep=2,fill=white,opacity=0.7,text opacity=1] (A) at (0.4,0.4) {\Large 11};\end{tikzpicture}}
\includegraphics[height=3.1 cm]{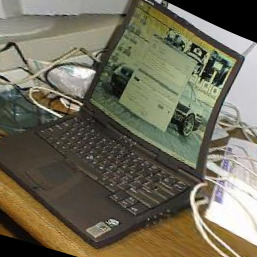} &	\includegraphics[height=3.1 cm]{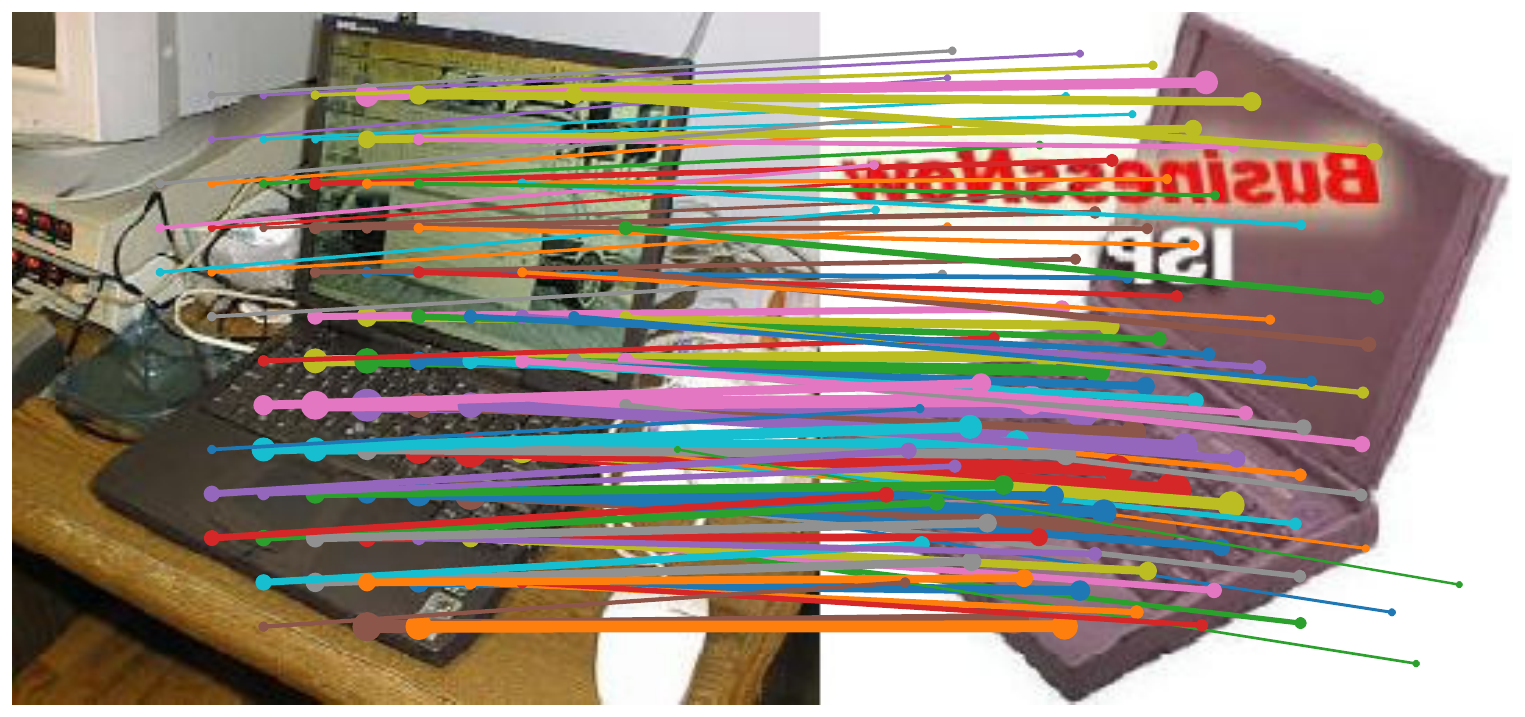} \\

\isArXiv{\includegraphics[height=2.7 cm]{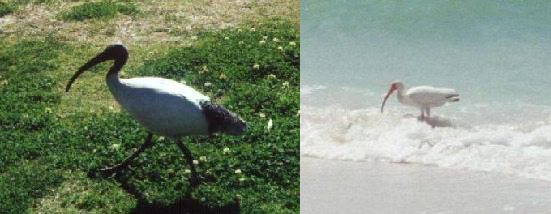}}{
\begin{tikzpicture}\node (label) at (0,0)[anchor=south west,inner sep=0,outer sep=0]{
\includegraphics[height=2.7 cm]{selected-caltech/774_pair}};\node [outer sep=1,inner sep=2,fill=white,opacity=0.7,text opacity=1] (A) at (0.4,0.4) {\Large 12};\end{tikzpicture}}
\includegraphics[height=2.7 cm]{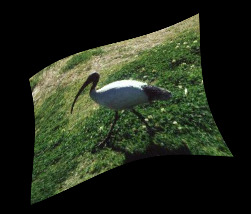} &	\includegraphics[height=2.7 cm]{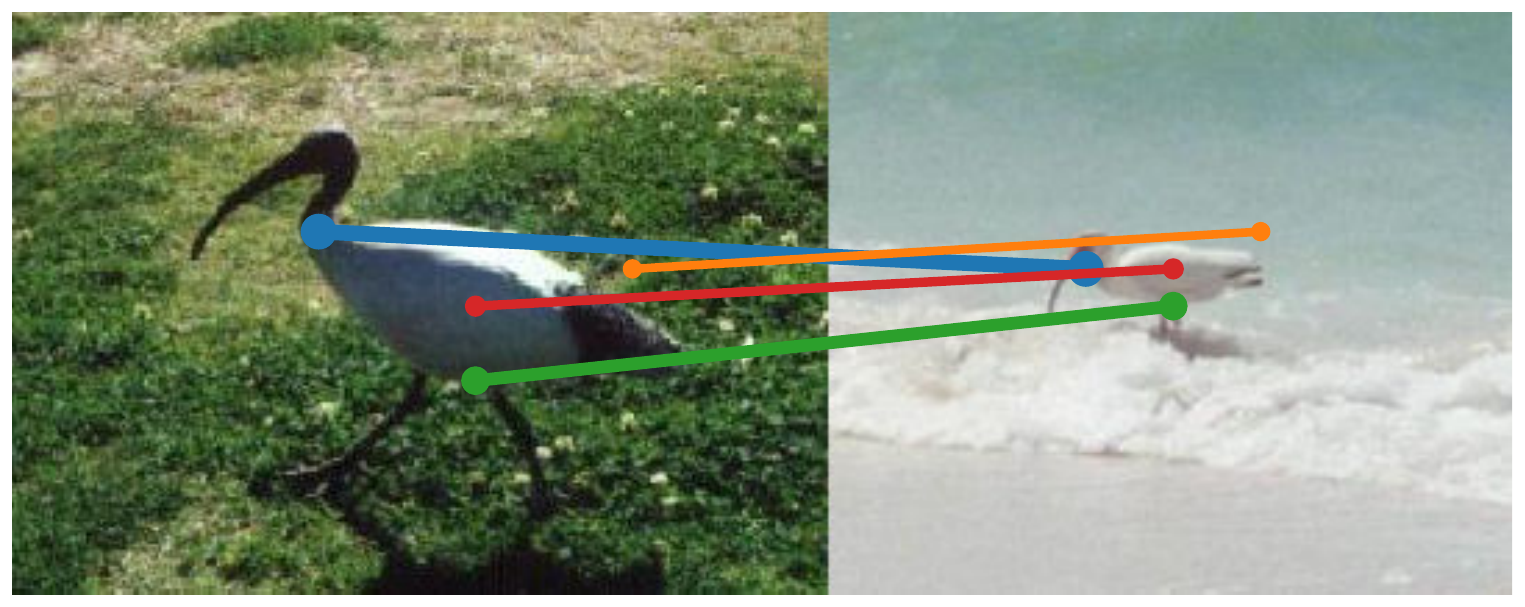} \\

\isArXiv{\includegraphics[height=2.8 cm]{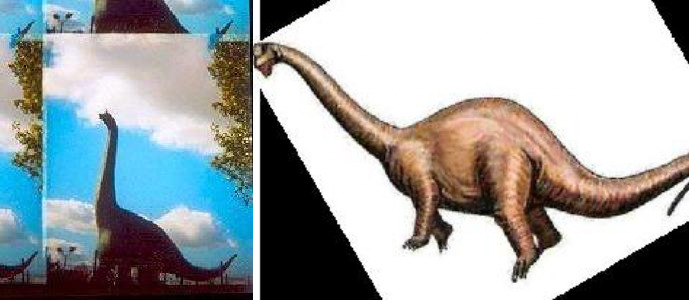}}{
\begin{tikzpicture}\node (label) at (0,0)[anchor=south west,inner sep=0,outer sep=0]{
\includegraphics[height=2.8 cm]{selected-caltech/217_pair}};\node [outer sep=1,inner sep=2,fill=white,opacity=0.7,text opacity=1] (A) at (0.4,0.4) {\Large 13};\end{tikzpicture}}
\includegraphics[height=2.8 cm]{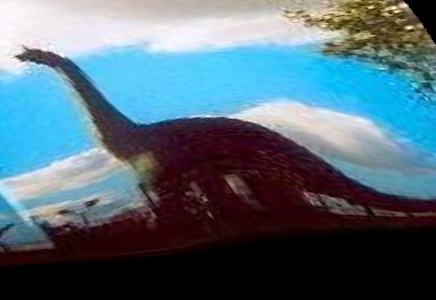} &	\includegraphics[height=2.8 cm]{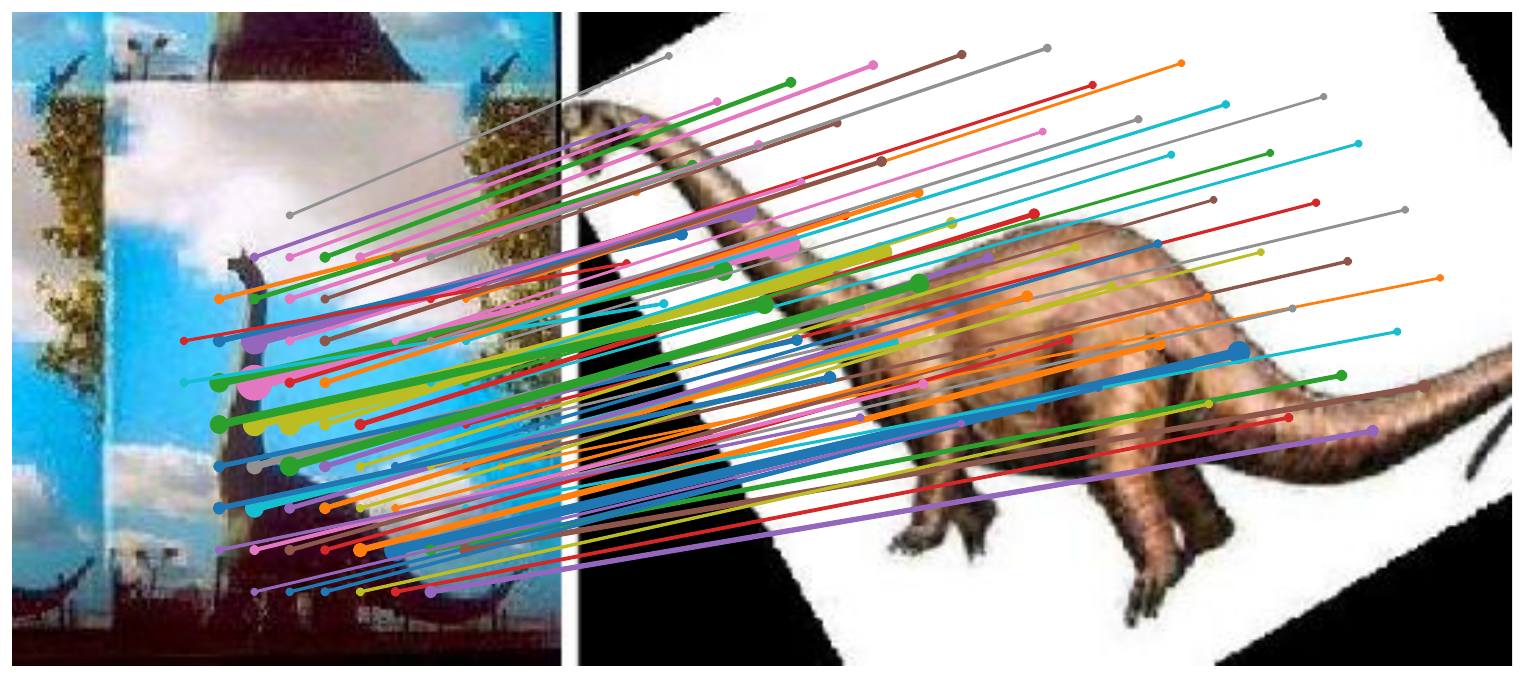} \\

\isArXiv{\includegraphics[height=2 cm]{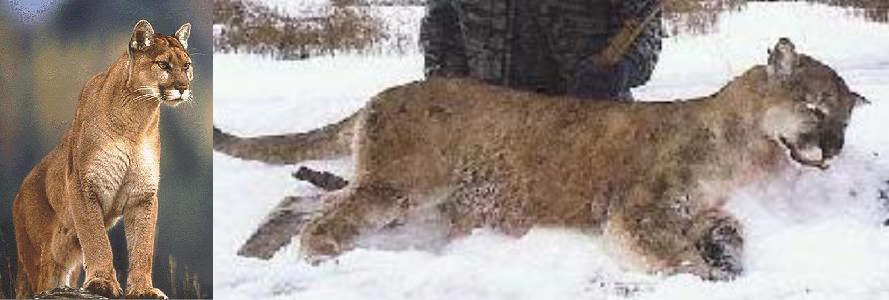}}{
\begin{tikzpicture}\node (label) at (0,0)[anchor=south west,inner sep=0,outer sep=0]{
\includegraphics[height=2 cm]{selected-caltech/369_pair}};\node [outer sep=1,inner sep=2,fill=white,opacity=0.7,text opacity=1] (A) at (0.4,0.4) {\Large 14};\end{tikzpicture}}
\includegraphics[height=2 cm]{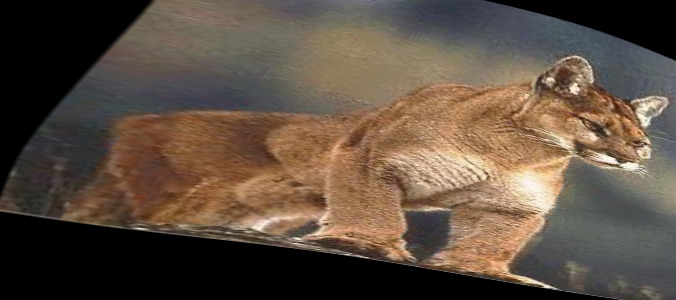} &	\includegraphics[height=2 cm]{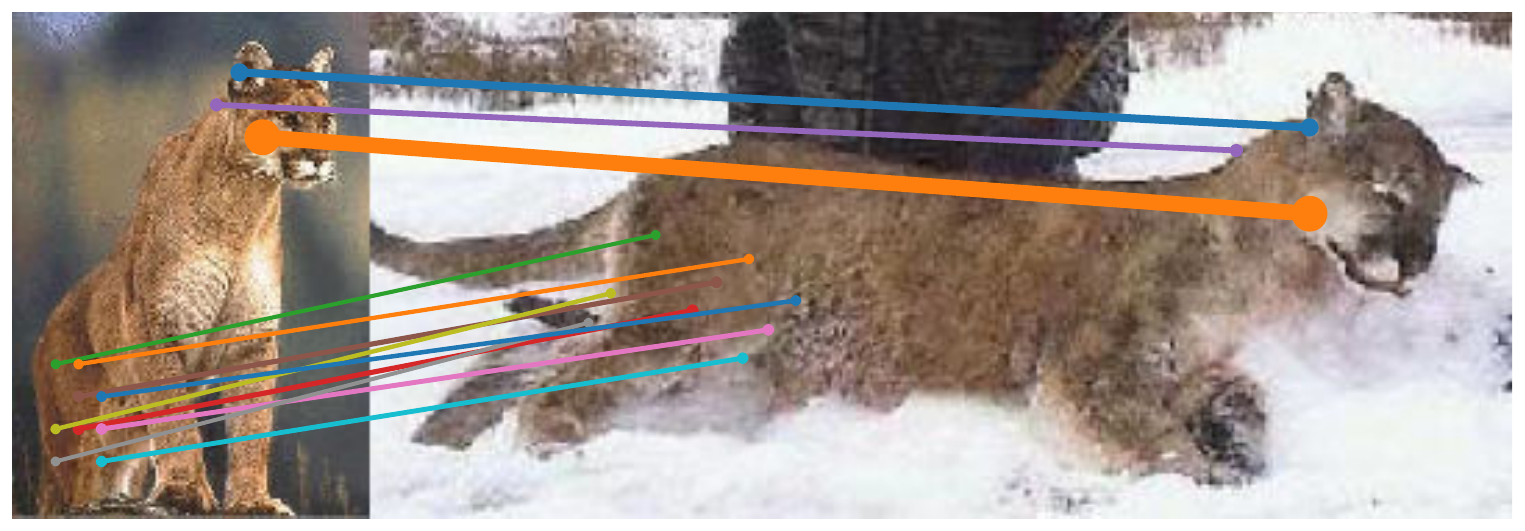} \\

(a) Semantic alignment & (b) Strongest inlier matches \\

\end{tabular}		
\captionsetup{font={small}}		
\caption{{\bf Additional examples on the Caltech-101 dataset.} Each row corresponds to one example. (a) shows the (right) automatic semantic alignment of the (left) source and (middle) target images. (b) shows the strongest inlier feature matches.}
\label{fig:qual_caltech_supp}		
\end{center}		
\end{figure*}

\begin{figure*}[t!]		
\begin{center}		
\setlength{\tabcolsep}{1pt} 
\renewcommand{\arraystretch}{1} 
\newcommand{\size}{2.6}	
\begin{tabular}{cc}		

\isArXiv{\includegraphics[height=2.6 cm]{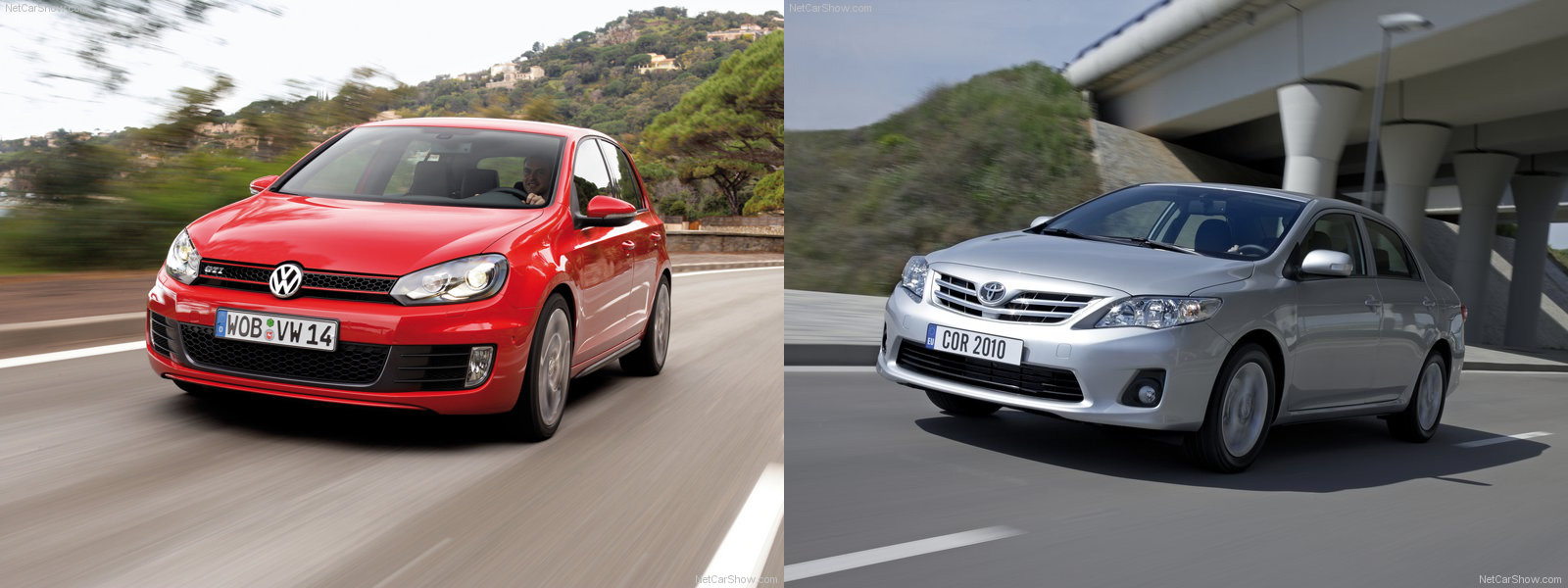}}{
\begin{tikzpicture}\node (label) at (0,0)[anchor=south west,inner sep=0,outer sep=0]{
\includegraphics[height=2.6 cm]{selected-tss/28_pair}};\node [outer sep=1,inner sep=2,fill=white,opacity=0.7,text opacity=1] (A) at (0.4,0.4) {\Large 15};\end{tikzpicture}}
\includegraphics[height=2.6 cm]{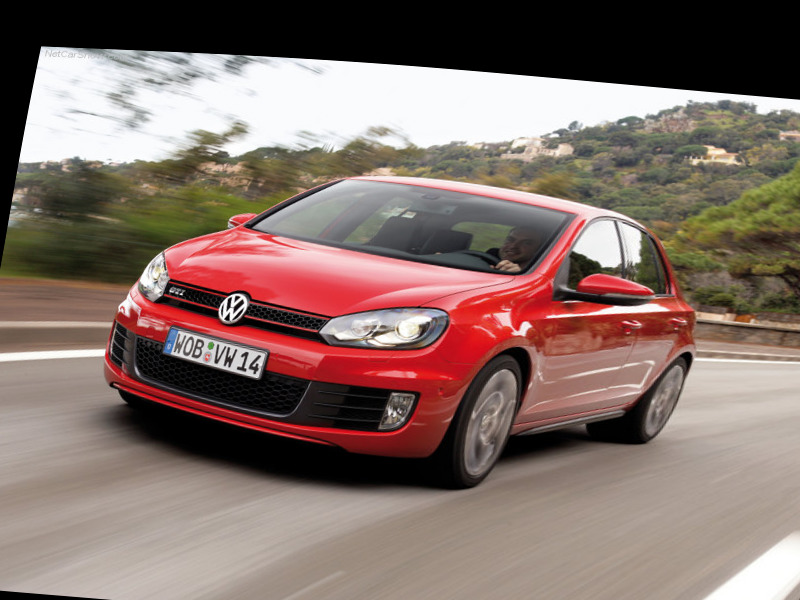} &	\includegraphics[height=2.6 cm]{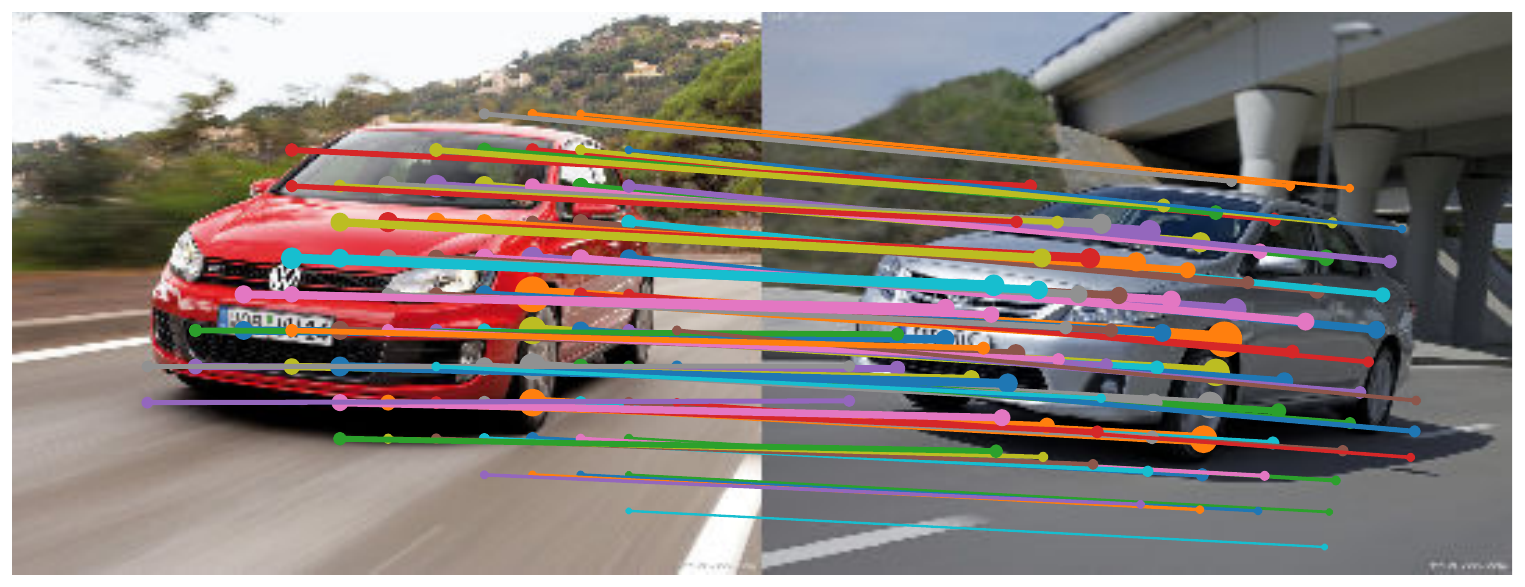} \\

\isArXiv{\includegraphics[height=2.6 cm]{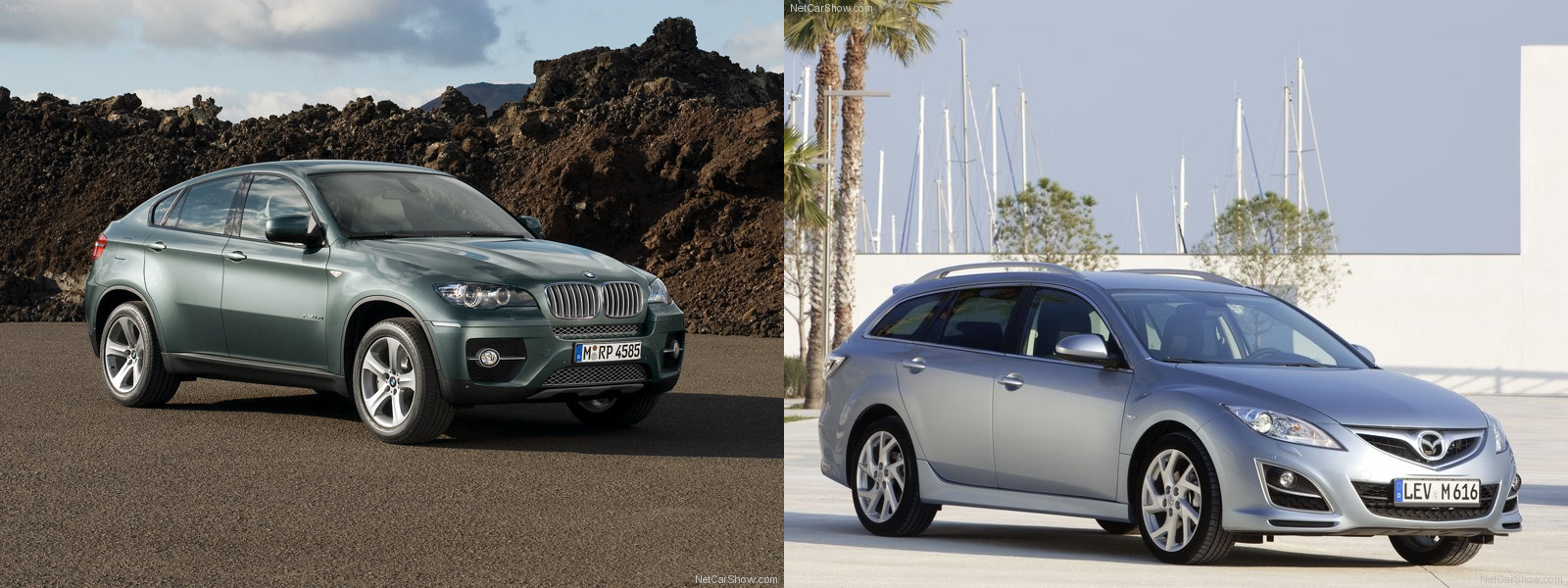}}{
\begin{tikzpicture}\node (label) at (0,0)[anchor=south west,inner sep=0,outer sep=0]{
\includegraphics[height=2.6 cm]{selected-tss/63_pair}};\node [outer sep=1,inner sep=2,fill=white,opacity=0.7,text opacity=1] (A) at (0.4,0.4) {\Large 16};\end{tikzpicture}}
\includegraphics[height=2.6 cm]{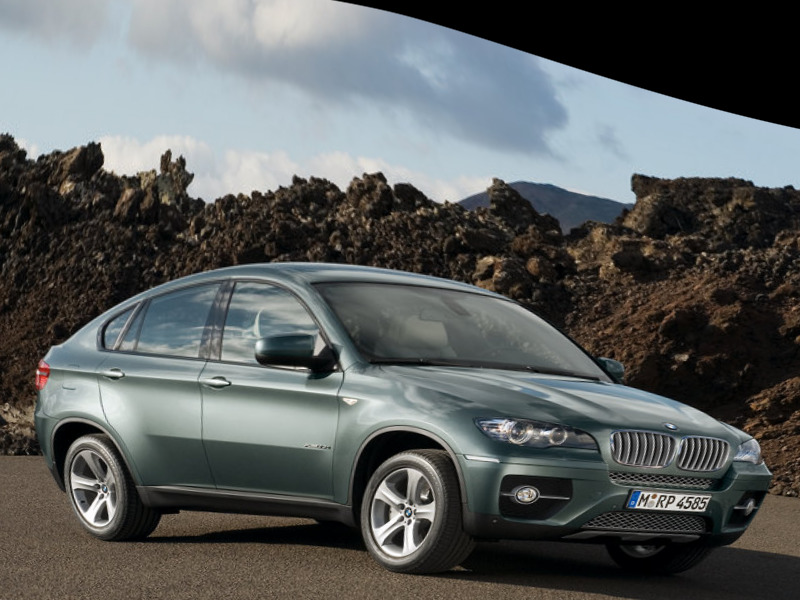} &	\includegraphics[height=2.6 cm]{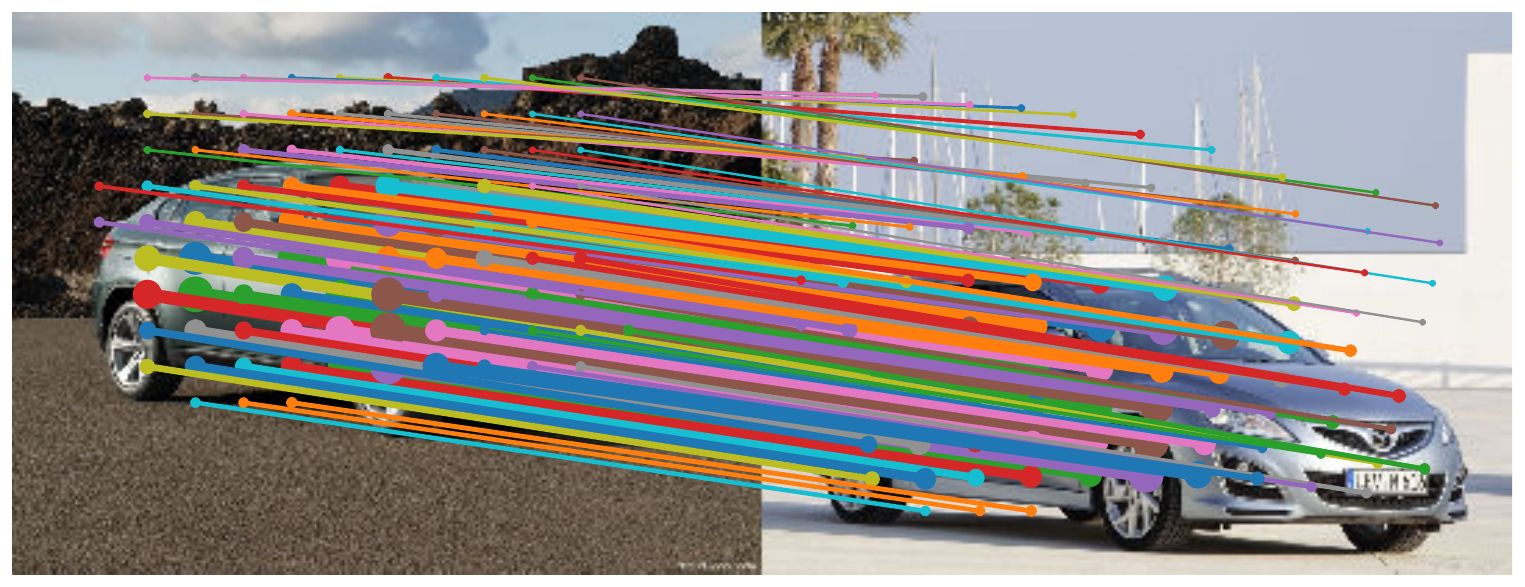} \\

\isArXiv{\includegraphics[height=2.6 cm]{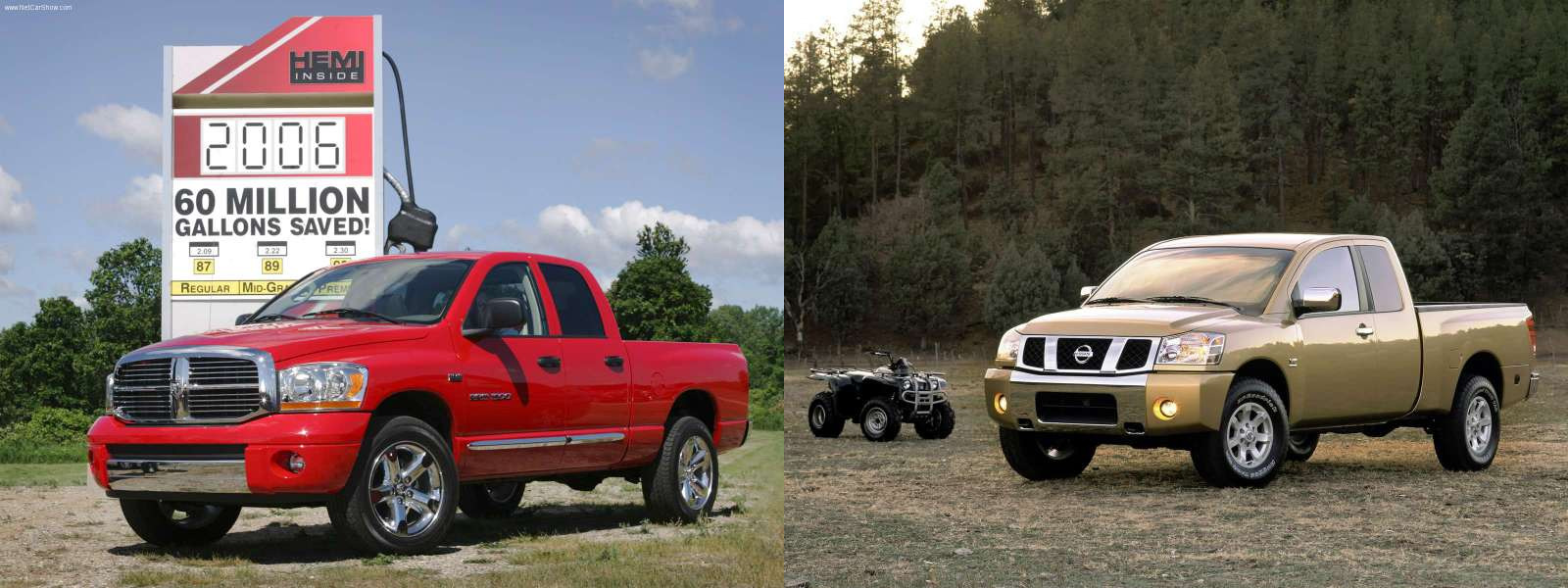}}{
\begin{tikzpicture}\node (label) at (0,0)[anchor=south west,inner sep=0,outer sep=0]{
\includegraphics[height=2.6 cm]{selected-tss/328_pair}};\node [outer sep=1,inner sep=2,fill=white,opacity=0.7,text opacity=1] (A) at (0.4,0.4) {\Large 17};\end{tikzpicture}}
\includegraphics[height=2.6 cm]{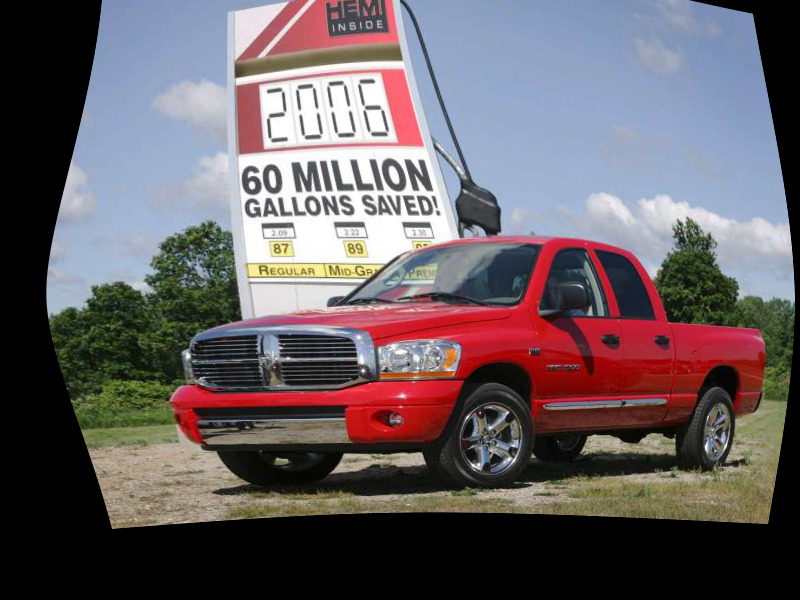} &	\includegraphics[height=2.6 cm]{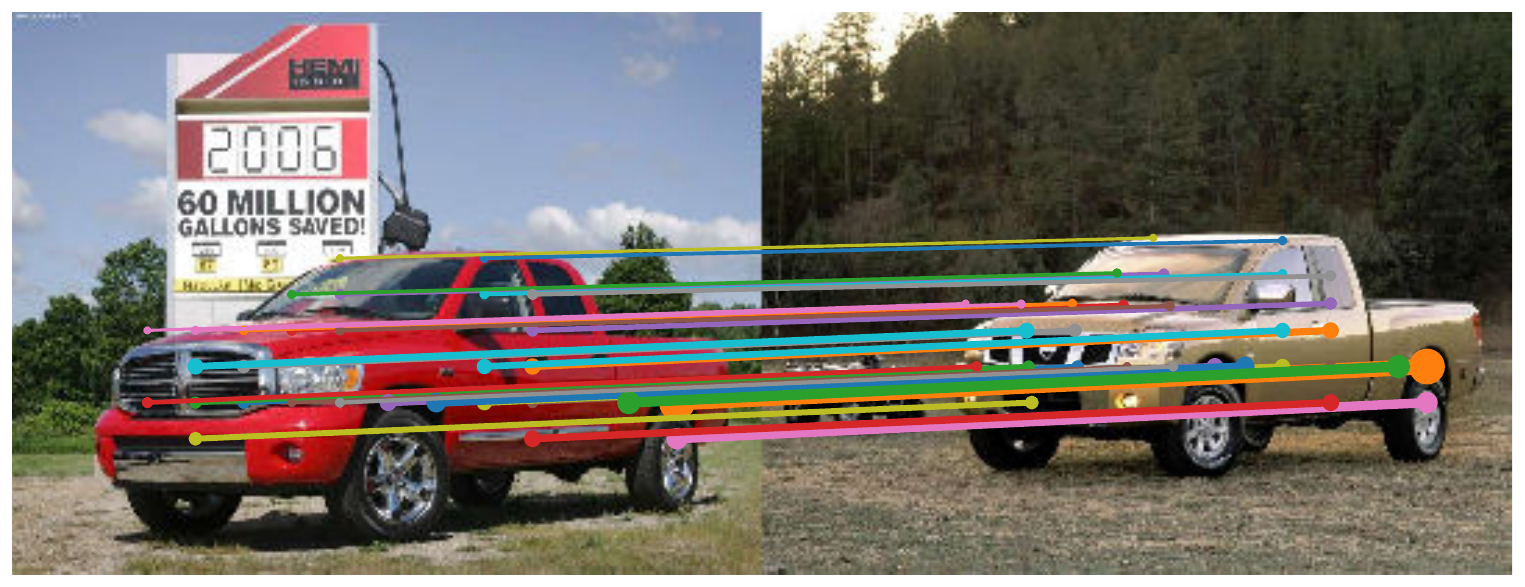} \\

\isArXiv{\includegraphics[height=2.9 cm]{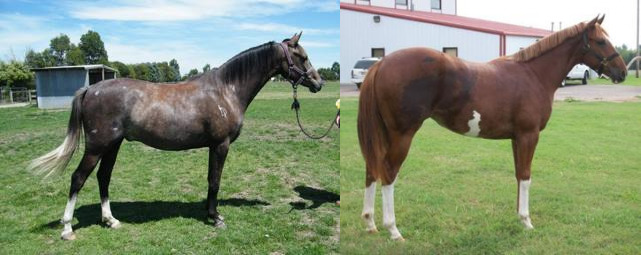}}{
\begin{tikzpicture}\node (label) at (0,0)[anchor=south west,inner sep=0,outer sep=0]{
\includegraphics[height=2.9 cm]{selected-tss/406_pair}};\node [outer sep=1,inner sep=2,fill=white,opacity=0.7,text opacity=1] (A) at (0.4,0.4) {\Large 18};\end{tikzpicture}}
\includegraphics[height=2.9 cm]{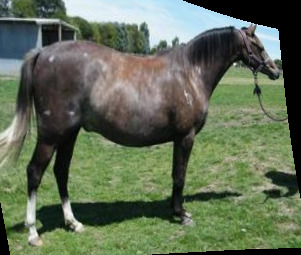} &	\includegraphics[height=2.9 cm]{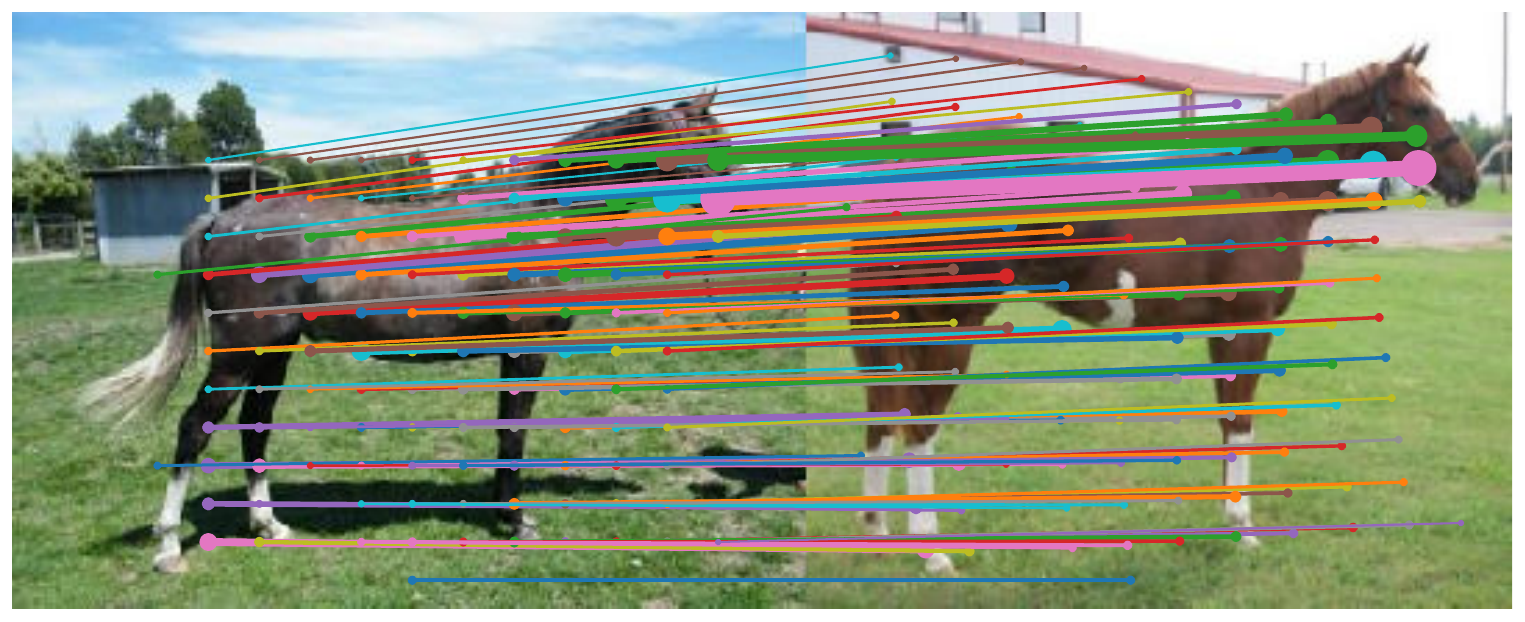} \\

\isArXiv{\includegraphics[height=2.6 cm]{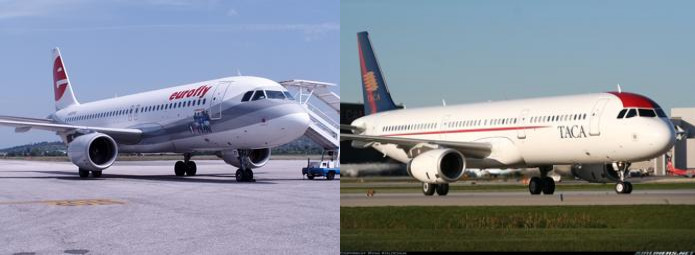}}{
\begin{tikzpicture}\node (label) at (0,0)[anchor=south west,inner sep=0,outer sep=0]{
\includegraphics[height=2.6 cm]{selected-tss/464_pair}};\node [outer sep=1,inner sep=2,fill=white,opacity=0.7,text opacity=1] (A) at (0.4,0.4) {\Large 19};\end{tikzpicture}}
\includegraphics[height=2.6 cm]{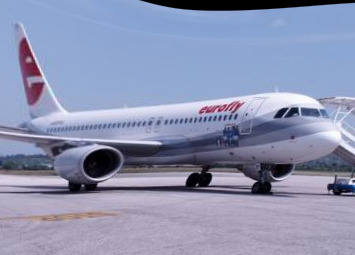} &	\includegraphics[height=2.6 cm]{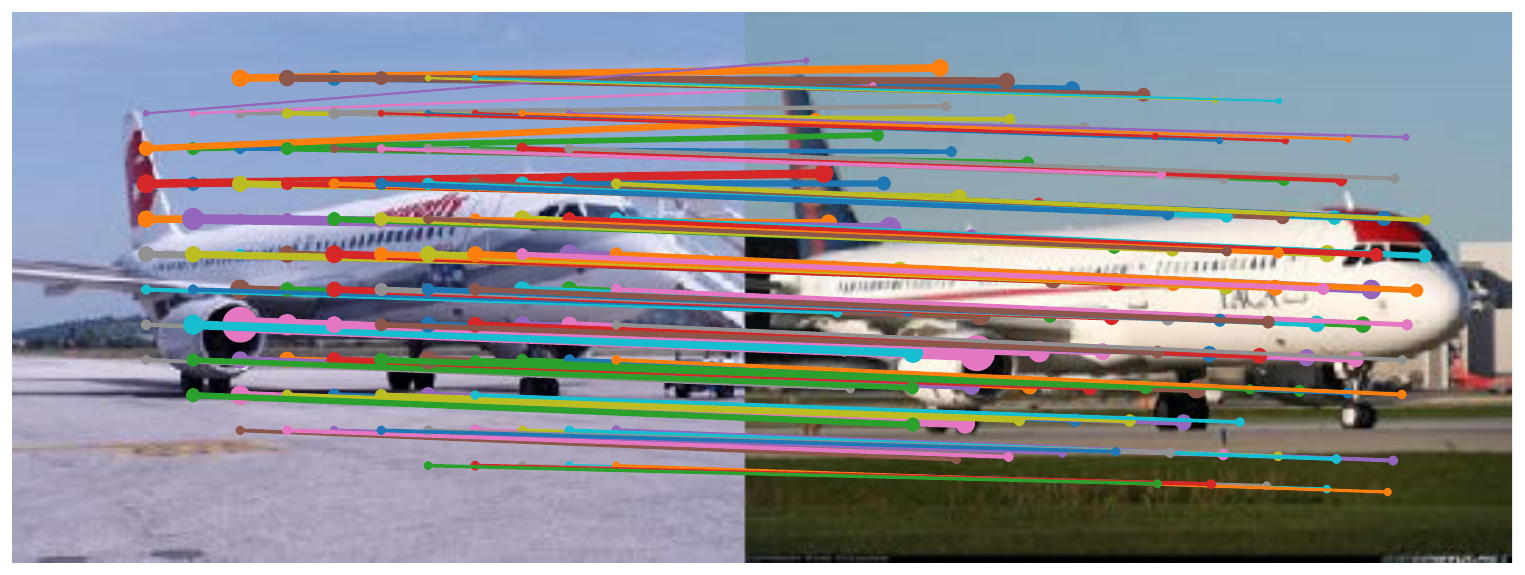} \\

\isArXiv{\includegraphics[height=2.6 cm]{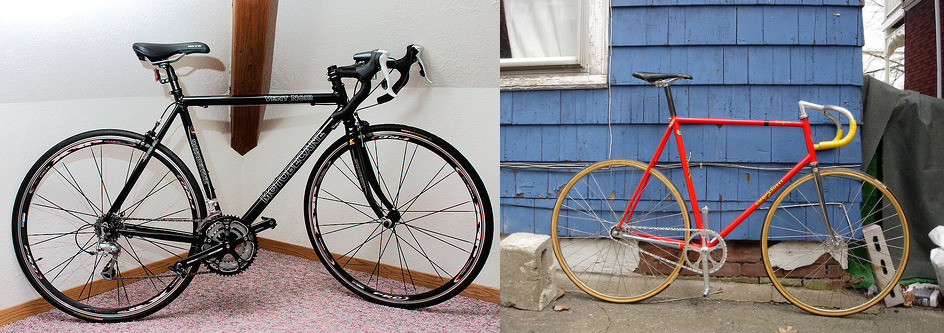}}{
\begin{tikzpicture}\node (label) at (0,0)[anchor=south west,inner sep=0,outer sep=0]{
\includegraphics[height=2.6 cm]{selected-tss/558_pair}};\node [outer sep=1,inner sep=2,fill=white,opacity=0.7,text opacity=1] (A) at (0.4,0.4) {\Large 20};\end{tikzpicture}}
\includegraphics[height=2.6 cm]{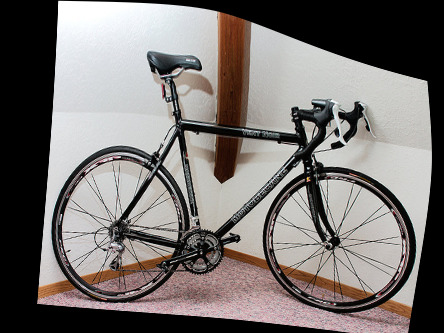} &	\includegraphics[height=2.6 cm]{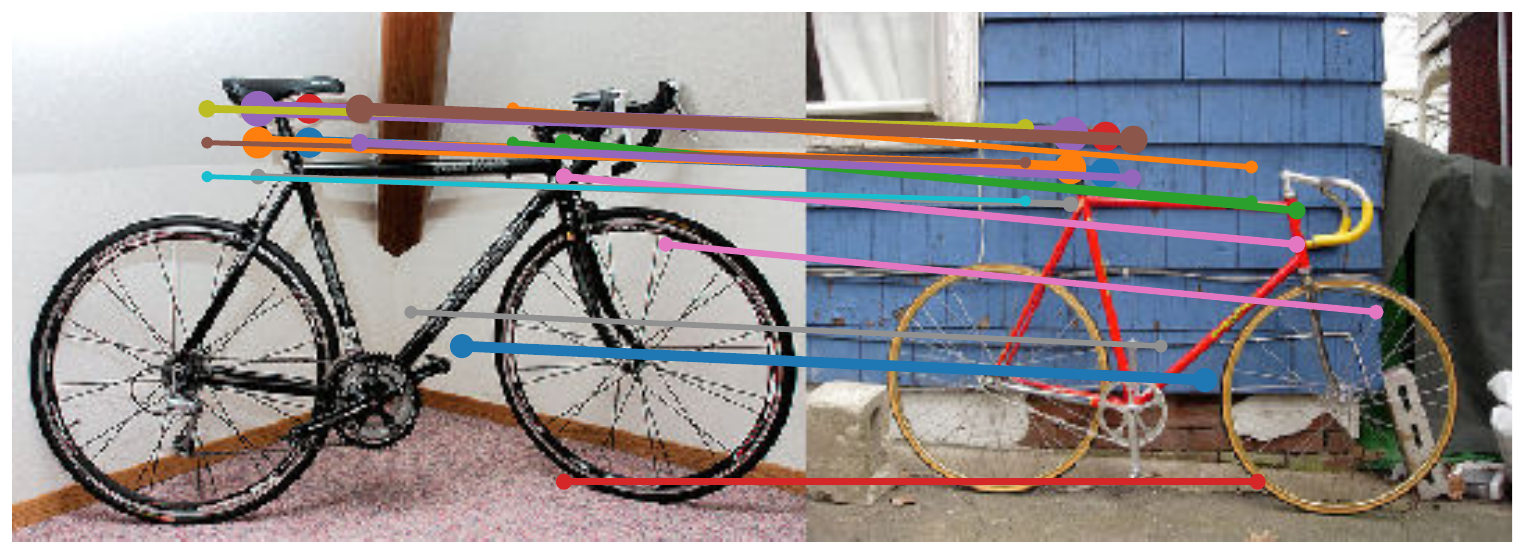} \\

\isArXiv{\includegraphics[height=2.6 cm]{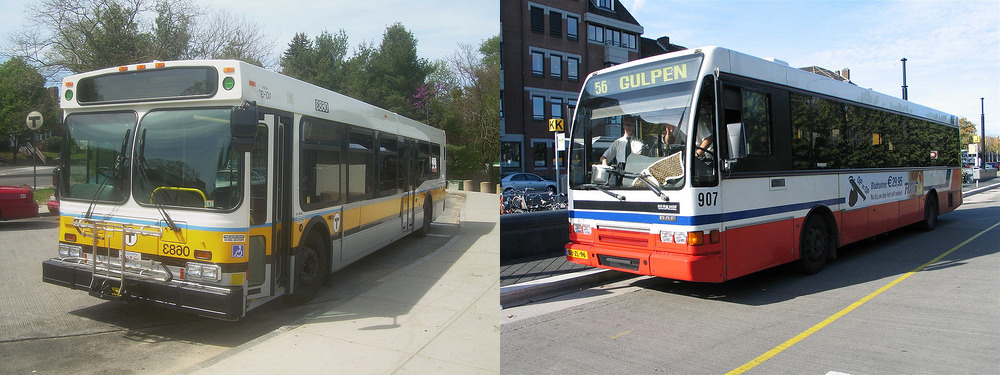}}{
\begin{tikzpicture}\node (label) at (0,0)[anchor=south west,inner sep=0,outer sep=0]{
\includegraphics[height=2.6 cm]{selected-tss/674_pair}};\node [outer sep=1,inner sep=2,fill=white,opacity=0.7,text opacity=1] (A) at (0.4,0.4) {\Large 21};\end{tikzpicture}}
\includegraphics[height=2.6 cm]{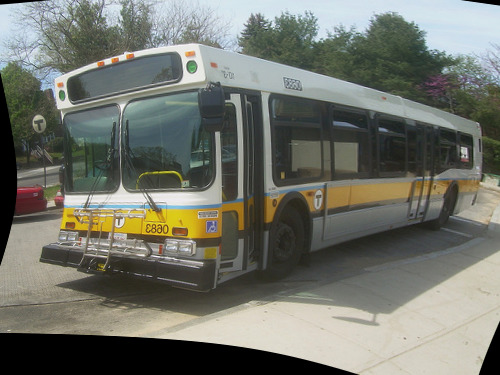} &	\includegraphics[height=2.6 cm]{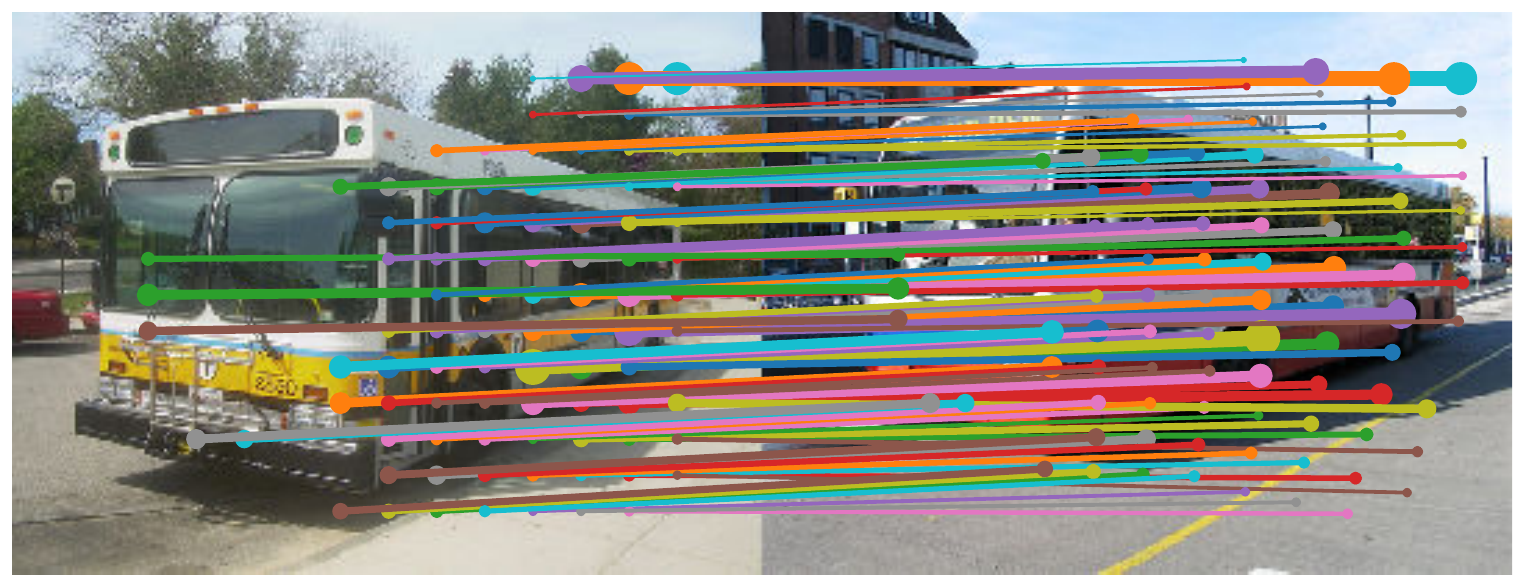} \\

(a) Semantic alignment & (b) Strongest inlier matches \\

\end{tabular}		
\captionsetup{font={small}}		
\caption{{\bf Additional examples on the TSS dataset.} Each row corresponds to one example. (a) shows the (right) automatic semantic alignment of the (left) source and (middle) target images. (b) shows the strongest inlier feature matches.}
\label{fig:qual_tss_supp}		
\end{center}		
\end{figure*}

\begin{figure*}[t!]		
\begin{center}		
\setlength{\tabcolsep}{1pt} 
\renewcommand{\arraystretch}{1} 
\newcommand{\size}{2.5}	
\begin{tabular}{cc}		

\isArXiv{\includegraphics[height=3.1 cm]{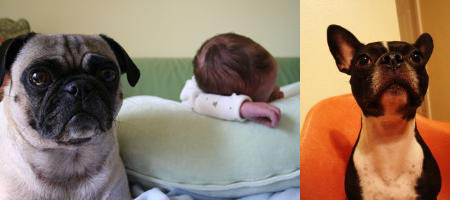}}{
\begin{tikzpicture}\node (label) at (0,0)[anchor=south west,inner sep=0,outer sep=0]{
        \includegraphics[height=3.1 cm]{selected-pf-pascal/173_pair}};\node [outer sep=1,inner sep=2,fill=white,opacity=0.7,text opacity=1] (A) at (0.4,0.4) {\Large 1};\end{tikzpicture}}
\includegraphics[height=3.1 cm]{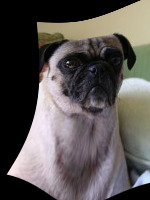} &	\includegraphics[height=3.1 cm]{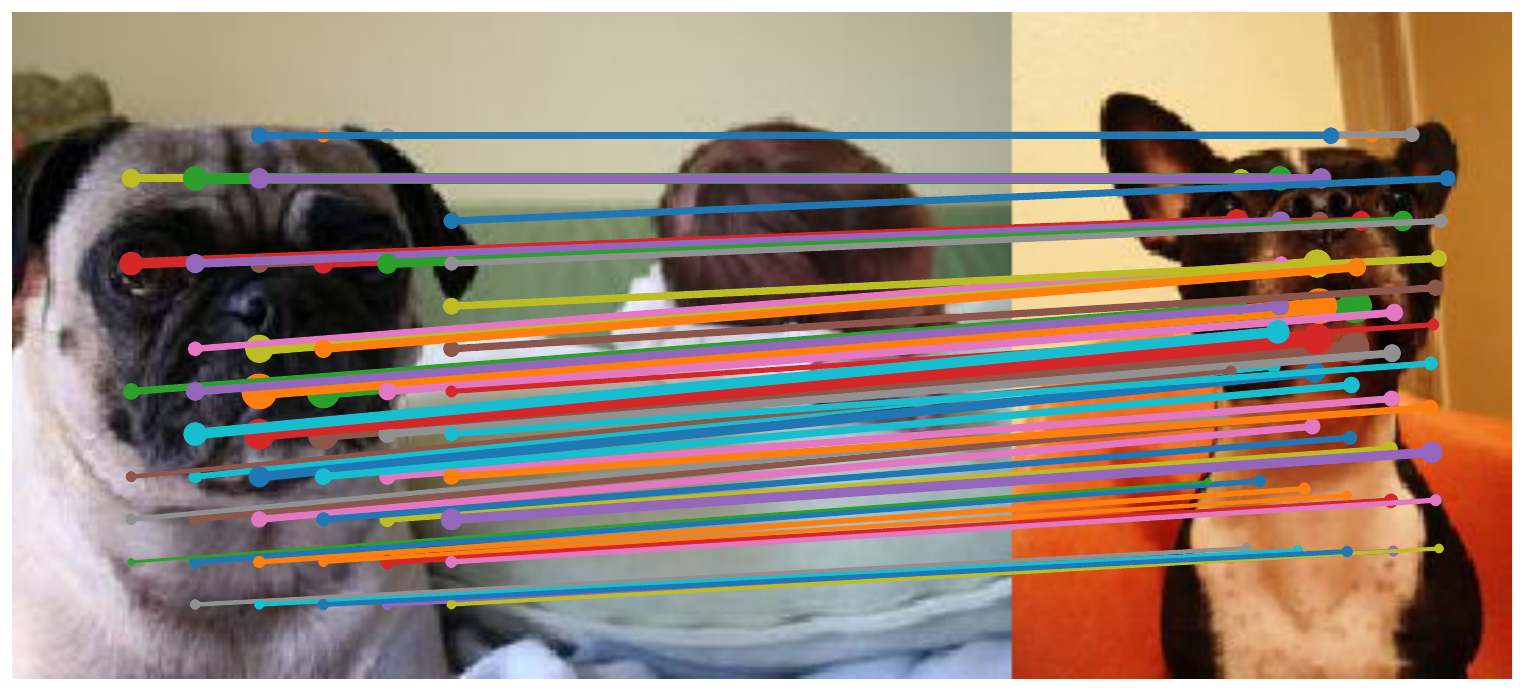} \\

\isArXiv{\includegraphics[height=3.1 cm]{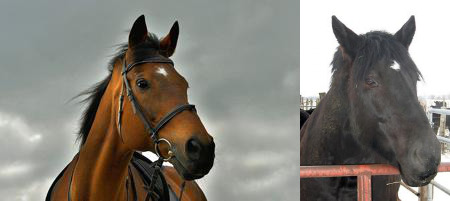}}{
\begin{tikzpicture}\node (label) at (0,0)[anchor=south west,inner sep=0,outer sep=0]{
\includegraphics[height=3.1 cm]{selected-pf-pascal/195_pair}};\node [outer sep=1,inner sep=2,fill=white,opacity=0.7,text opacity=1] (A) at (0.4,0.4) {\Large 2};\end{tikzpicture}}
\includegraphics[height=3.1 cm]{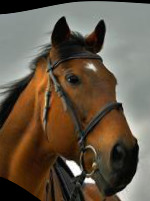} &	\includegraphics[height=3.1 cm]{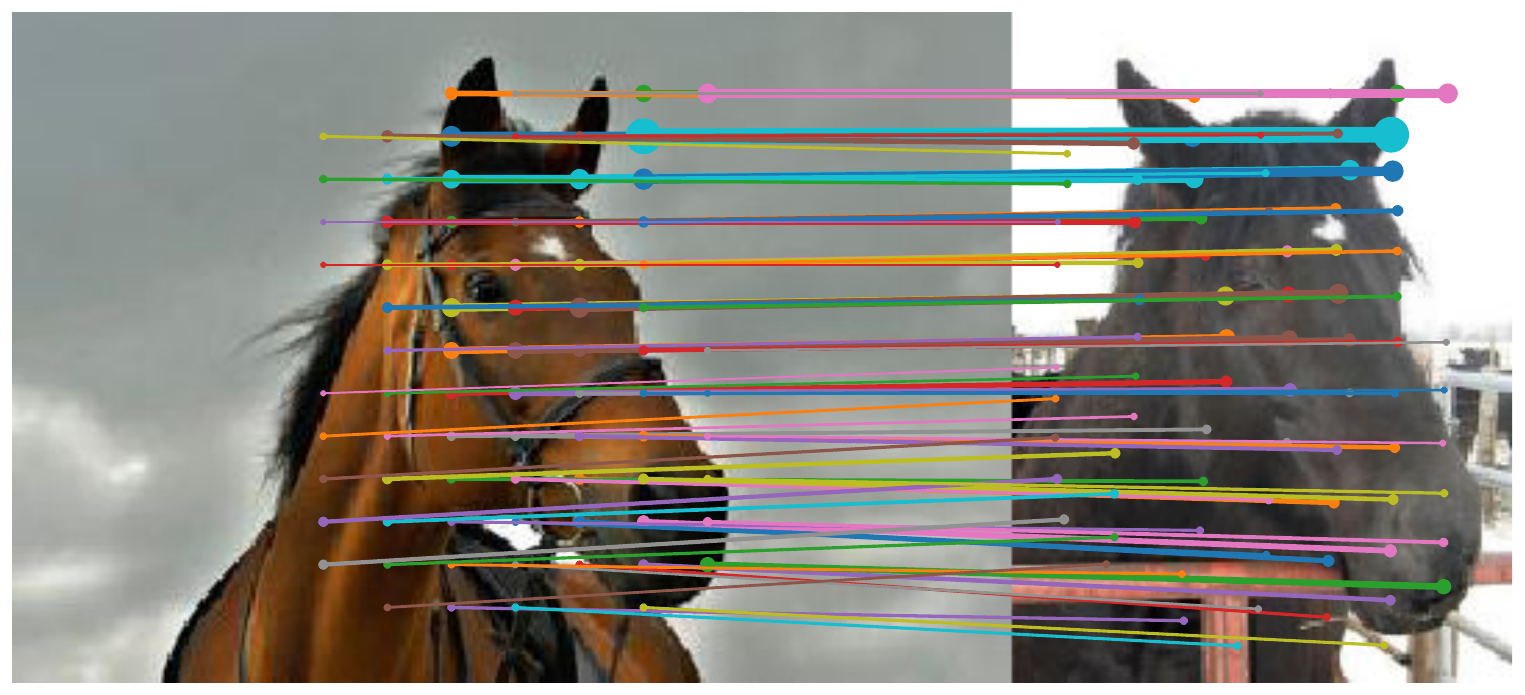} \\

\isArXiv{\includegraphics[height=3.1 cm]{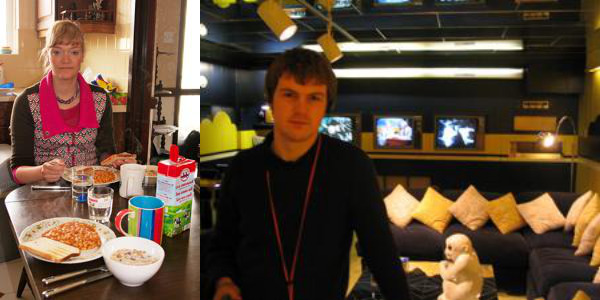}}{
\begin{tikzpicture}\node (label) at (0,0)[anchor=south west,inner sep=0,outer sep=0]{
\includegraphics[height=3.1 cm]{selected-pf-pascal/241_pair}};\node [outer sep=1,inner sep=2,fill=white,opacity=0.7,text opacity=1] (A) at (0.4,0.4) {\Large 3};\end{tikzpicture}}
\includegraphics[height=3.1 cm]{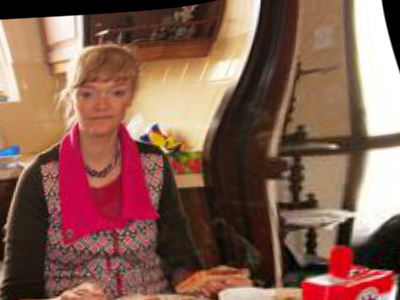} &	\includegraphics[height=3.1 cm]{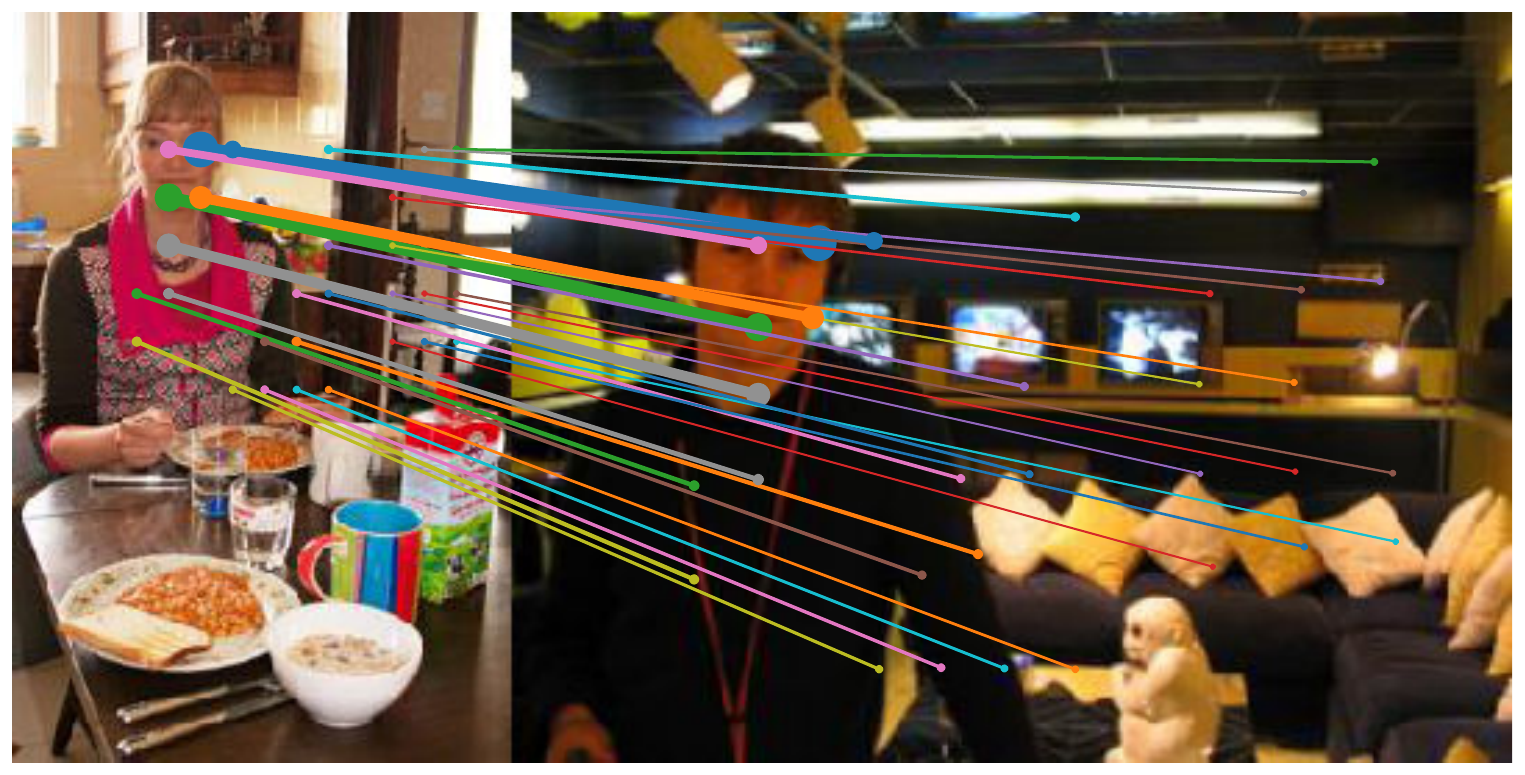} \\

\isArXiv{\includegraphics[height=2.6 cm]{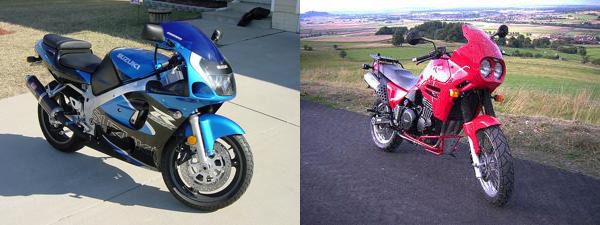}}{
\begin{tikzpicture}\node (label) at (0,0)[anchor=south west,inner sep=0,outer sep=0]{
\includegraphics[height=2.6 cm]{selected-pf-pascal/212_pair}};\node [outer sep=1,inner sep=2,fill=white,opacity=0.7,text opacity=1] (A) at (0.4,0.4) {\Large 4};\end{tikzpicture}}
\includegraphics[height=2.6 cm]{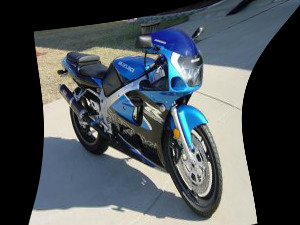} &	\includegraphics[height=2.6 cm]{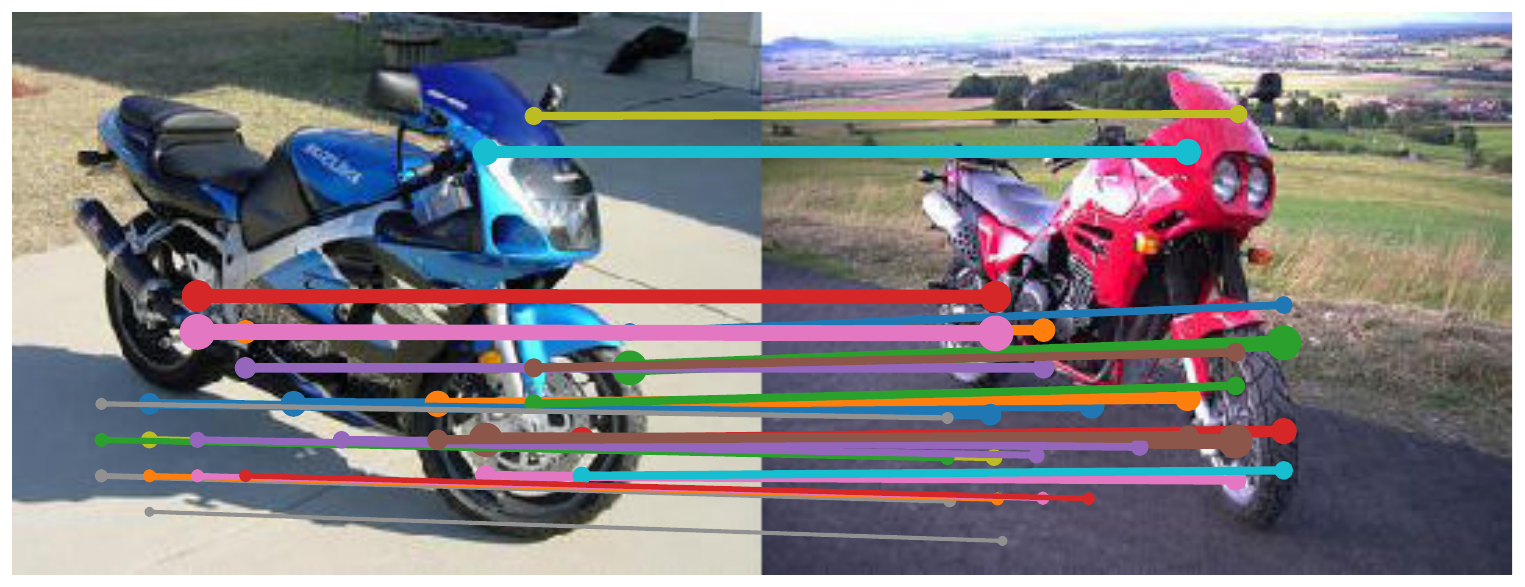} \\

\isArXiv{\includegraphics[height=2.6 cm]{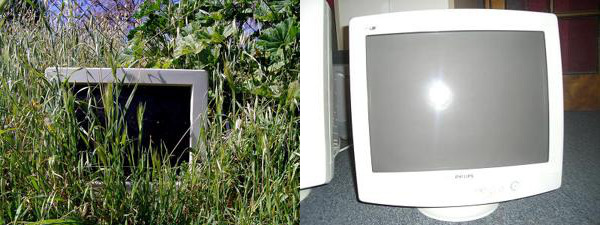}}{
\begin{tikzpicture}\node (label) at (0,0)[anchor=south west,inner sep=0,outer sep=0]{
\includegraphics[height=2.6 cm]{selected-pf-pascal/288_pair}};\node [outer sep=1,inner sep=2,fill=white,opacity=0.7,text opacity=1] (A) at (0.4,0.4) {\Large 5};\end{tikzpicture}}
\includegraphics[height=2.6 cm]{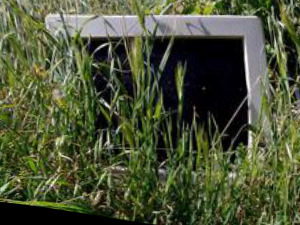} &	\includegraphics[height=2.6 cm]{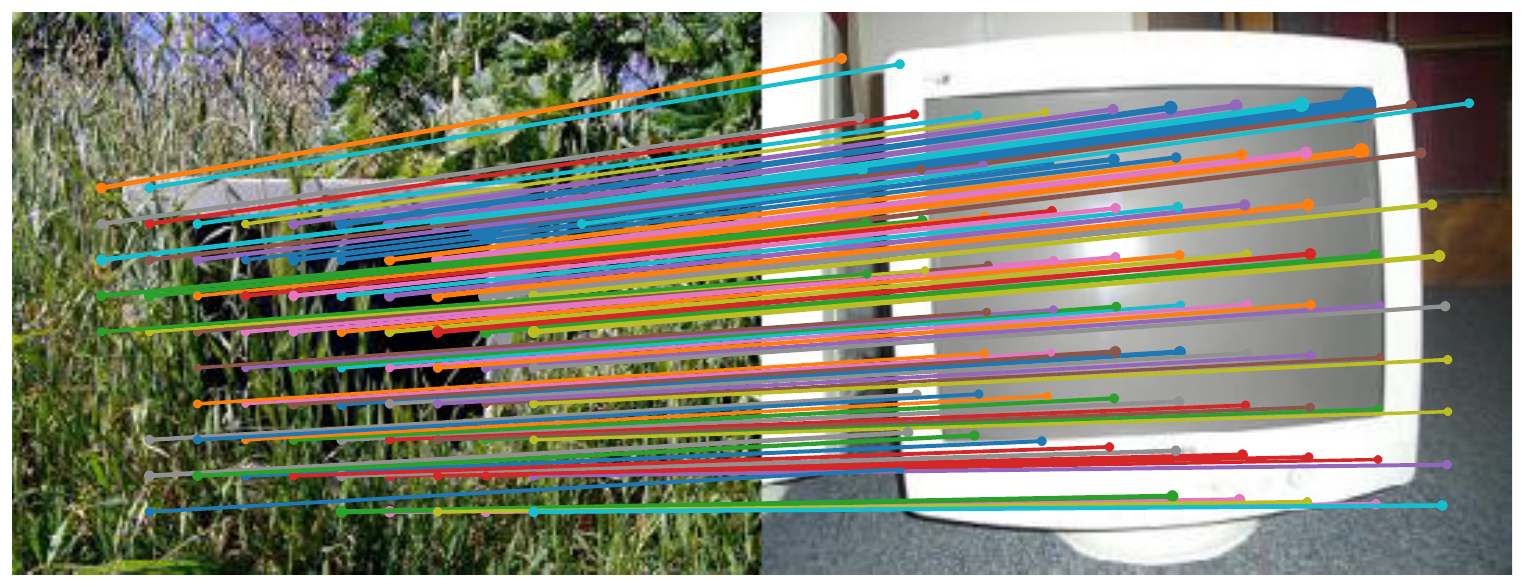} \\

\isArXiv{\includegraphics[height=2.5 cm]{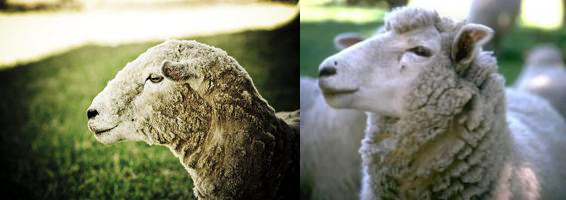}}{
\begin{tikzpicture}\node (label) at (0,0)[anchor=south west,inner sep=0,outer sep=0]{
\includegraphics[height=2.5 cm]{selected-pf-pascal/250_pair}};\node [outer sep=1,inner sep=2,fill=white,opacity=0.7,text opacity=1] (A) at (0.4,0.4) {\Large 6};\end{tikzpicture}}
\includegraphics[height=2.5 cm]{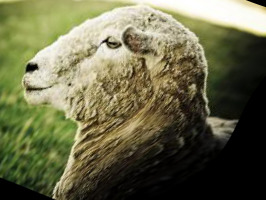} &	\includegraphics[height=2.5 cm]{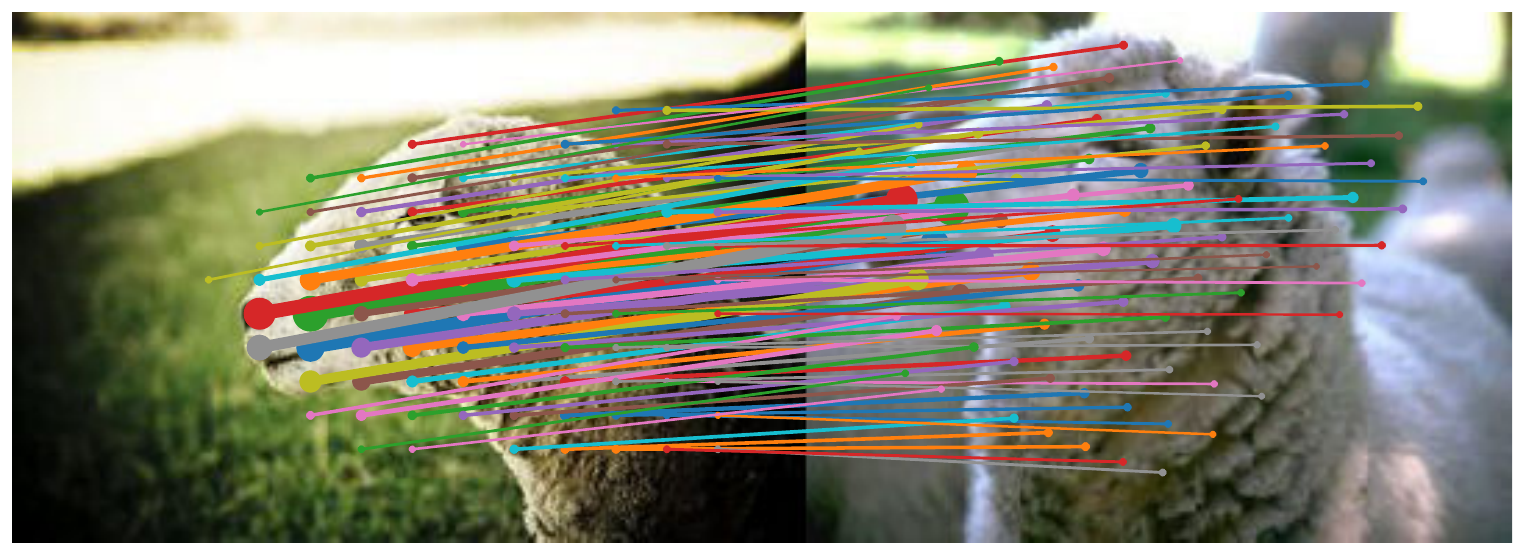} \\

\isArXiv{\includegraphics[height=2.5 cm]{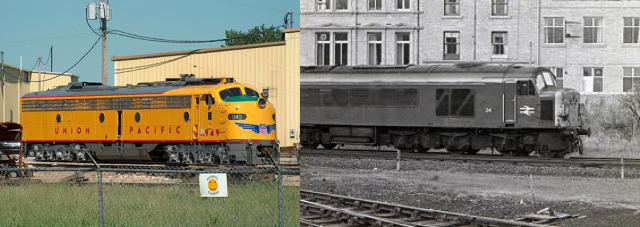}}{
\begin{tikzpicture}\node (label) at (0,0)[anchor=south west,inner sep=0,outer sep=0]{
\includegraphics[height=2.5 cm]{selected-pf-pascal/267_pair}};\node [outer sep=1,inner sep=2,fill=white,opacity=0.7,text opacity=1] (A) at (0.4,0.4) {\Large 7};\end{tikzpicture}}
\includegraphics[height=2.5 cm]{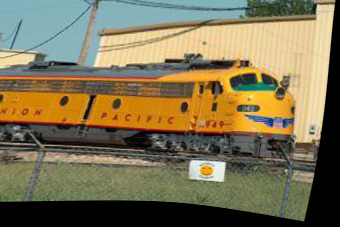} &	\includegraphics[height=2.5 cm]{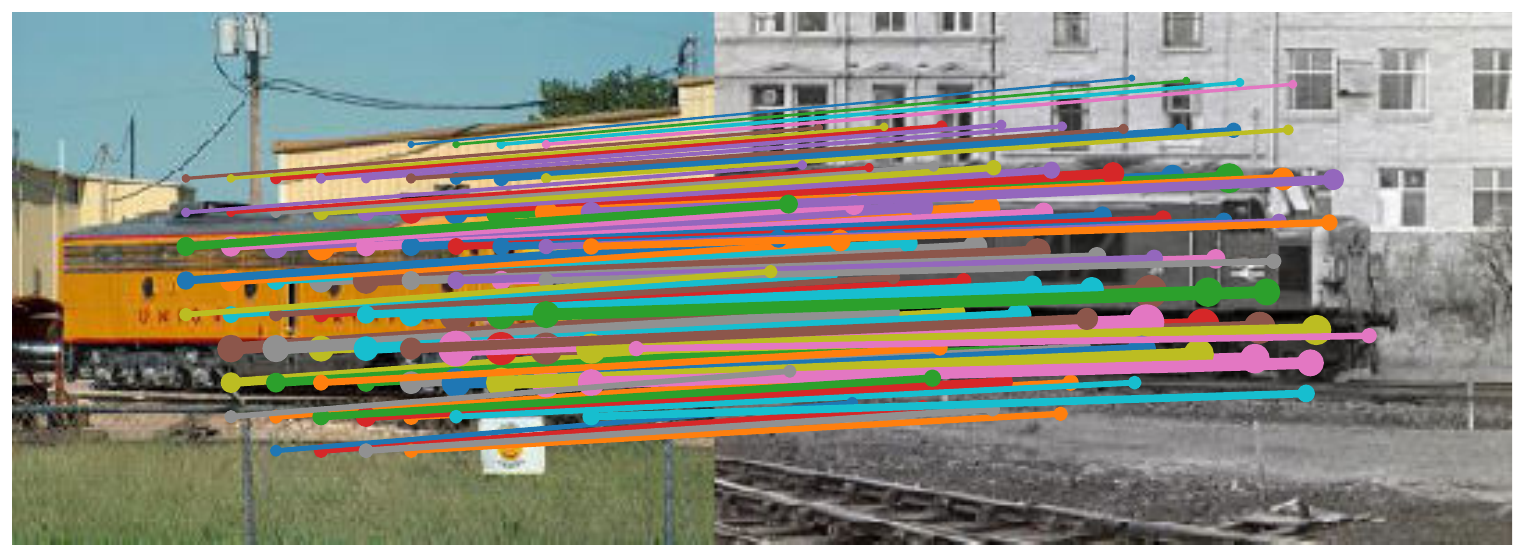} \\

(a) Semantic alignment & (b) Strongest inlier matches \\

\end{tabular}		
\captionsetup{font={small}}		
\caption{{\bf Additional examples on the PF-PASCAL dataset.} Each row corresponds to one example. (a) shows the (right) automatic semantic alignment of the (left) source and (middle) target images. (b) shows the strongest inlier feature matches.}
\label{fig:qual_pf_pascal_supp}		
\end{center}		
\end{figure*}

\paragraph{Qualitative results.} 
Figures~\ref{fig:qual_caltech} and~\ref{fig:qual_caltech_supp} show qualitative results on the Caltech-101 dataset, figures~\ref{fig:qual_tss} and~\ref{fig:qual_tss_supp} on the TSS dataset, and figures~\ref{fig:qual_pf_pascal} and~\ref{fig:qual_pf_pascal_supp} on the PF-PASCAL dataset.
Our method is able to align images across prominent viewpoint changes,
in the presence of significant clutter,
while simultaneously tolerating large intra-class variations. 

\begin{table}[t!]
\small
\begin{center}
\begin{tabular}{l@{\hskip 2mm}c@{\hskip 2mm}c@{\hskip 2mm}c@{\hskip 2mm}c}
\toprule					
Method &	FG3D. &	JODS &	PASC. &	avg.	\\ \midrule \midrule
HOG+PF-LOM~\cite{ham2017} &	0.786 &	0.653 &	0.531 &	0.657	\\
HOG+TSS~\cite{taniai2016joint} &	0.830 &	0.595 &	0.483 &	0.636	\\
FCSS+SIFT Flow~\cite{kim2017fcss} &	0.830 &	0.656 &	0.494 &	0.660	\\
FCSS+PF-LOM~\cite{kim2017fcss} &	0.839 &	0.635 &	0.582 &	0.685	\\
HOG+OADSC~\cite{yang2017object} &	0.875 &	0.708 &	{\bf 0.729} &	{\bf 0.771}	\\
FCSS+DCTM~\cite{kim2017dctm} &	0.891 &	0.721 &	0.610 &	0.740	\\
VGG-16+CNNGeo~\cite{Rocco17} &	0.839 &	0.658 &	0.528 &	0.675	\\
ResNet-101+CNNGeo~\cite{Rocco17} &	0.901 &	{\bf 0.764} &	0.563 &	0.743	\\
Proposed &	{\bf 0.903} &	{\bf 0.764} &	0.565 &	0.744	\\
\bottomrule
\end{tabular}
\captionsetup{font={small}}
\vspace{-1mm}
\captionof{table}{{\bf Evaluation results on the TSS dataset.}}
\label{tab:eval_tss}
\vspace{-7mm}
\end{center}
\end{table}					

\section{Conclusions}
We have designed a network architecture and training procedure
for semantic image alignment inspired by the robust inlier scoring used in the widely successful RANSAC fitting algorithm~\cite{fischler1981random}. The architecture requires supervision only in the form of matching 
image pairs
and sets the new state-of-the-art on multiple standard semantic alignment benchmarks, even beating alignment methods that require geometric supervision at training time. {However, handling multiple objects and non-matching image pairs still remains an open challenge.}
These results open-up the possibility of learning powerful correspondence networks from large-scale datasets such as ImageNet.

{\small
\paragraph{Acknowledgements.} 
This work has been partly supported by ERC grant LEAP (no.\ 336845), the Inria CityLab IPL, CIFAR Learning in Machines $\&$ Brains program and the European Regional Development Fund under the project IMPACT (reg. no.\ CZ$.02.1.01/0.0/0.0/15\_003/0000468$).
}

\FloatBarrier

{\small
\bibliographystyle{ieee}
\bibliography{egbib}
}


\end{document}